\documentclass[nonblindrev]{informs2017}
\OneAndAHalfSpacedXI 

\usepackage{endnotes}
\let\footnote=\endnote
\let\enotesize=\normalsize
\def\notesname{Endnotes}%
\def\makeenmark{\hbox to1.275em{\theenmark.\enskip\hss}}
\def\enoteformat{\rightskip0pt\leftskip0pt\parindent=1.275em
\leavevmode\llap{\makeenmark}}

\usepackage{graphicx}
\usepackage{subfigure}
\usepackage{multirow}
\usepackage{float}
\usepackage{epstopdf}
\usepackage{appendix}
\usepackage{bm}
\usepackage{dsfont}
\usepackage{appendix}
\usepackage{booktabs}
\usepackage{tablefootnote}
\usepackage[ruled]{algorithm2e}
\usepackage[para,online,flushleft]{threeparttable}

\newcommand{\tabincell}[2]{\begin{tabular}{@{}#1@{}}#2\end{tabular}}

\newcommand{\bmt}[1]{\tilde{\bm{#1}}}

\usepackage{color}
\usepackage[usenames,dvipsnames]{xcolor}
\definecolor{strcolor}{rgb}{0.6, 0.2, 0.6}
\definecolor{commentcolor}{rgb}{0.3125, 0.5, 0.3125}
\definecolor{keycol}{rgb}{0, 0, 1}

\usepackage{listings}
\lstdefinelanguage{Julia}%
{morekeywords={abstract,break,case,catch,const,continue,do,else,elseif,%
end,export,false,for,function,immutable,import,importall,if,in,%
macro,module,otherwise,quote,return,switch,true,try,type,typealias,%
using,while},%
sensitive=true,%
alsoother={\$},%
morecomment=[l]\#,%
morecomment=[n]{\#=}{=\#},%
morestring=[s]{"}{"},%
morestring=[m]{'}{'},%
}[keywords,comments,strings]%

\usepackage{natbib}
\bibpunct[, ]{(}{)}{,}{a}{}{,}%
\def\bibfont{\small}%
\TheoremsNumberedThrough     
\ECRepeatTheorems

\EquationsNumberedThrough    

\newcommand{\rr}[1]{{\color{black} #1}}
\newcommand{\newmodel}[1]{{\color{black} #1}}
\newcommand{\alg}[1]{{\color{black} #1}}
\newcommand{\newcontent}[1]{{\color{black} #1}}

\SetKw{Break}{break}
\SetKw{And}{and}
\SetKw{Input}{Input: }
\SetKw{Output}{Output: }

\begin{document}


\RUNAUTHOR{Ruan, Chen, Ho}

\RUNTITLE{Risk-Averse MDPs under Reward Ambiguity}

\TITLE{Risk-Averse MDPs under Reward Ambiguity}

\ARTICLEAUTHORS{%
\AUTHOR{Haolin Ruan}
\AFF{School of Data Science, City University of Hong Kong, Kowloon Tong, Hong Kong \\ haolin.ruan@my.cityu.edu.hk} 
\AUTHOR{Zhi Chen}
\AFF{Department of Management Sciences, College of Business, City University of Hong Kong, Kowloon Tong, Hong Kong \\ zhi.chen@cityu.edu.hk}
\AUTHOR{Chin Pang Ho}
\AFF{School of Data Science, City University of Hong Kong, Kowloon Tong, Hong Kong \\ clint.ho@cityu.edu.hk}
} 

\ABSTRACT{%
We propose a distributionally robust return-risk model for Markov decision processes (MDPs) under risk and reward ambiguity. The proposed model optimizes the weighted average of mean and percentile performances, and it covers the distributionally robust MDPs and the distributionally robust chance-constrained MDPs (both under reward ambiguity) as special cases. By considering that the unknown reward distribution lies in a Wasserstein ambiguity set, we derive the tractable reformulation for our model. In particular, we show that \newcontent{that the return-risk model can also account for risk from uncertain transition kernel when one only seeks deterministic policies,} and that a distributionally robust MDP under the percentile criterion can be reformulated as its nominal counterpart at an adjusted risk level. A scalable first-order algorithm is 
designed to solve large-scale problems, and we demonstrate the advantages of our proposed model and algorithm through numerical experiments. 
}%



\maketitle

%


\section{Introduction}

Markov decision processes (MDPs) provide a powerful modeling framework for sequential decision-making problems and reinforcement learning in stochastic dynamic environments \citep{puterman2014markov}. Obtaining the model parameters of MDPs that perfectly reflect the environments, however, has always been a challenge in practice, as these parameters are estimated from limited data that are potentially contaminated \citep{mannor2007bias}. Moreover, these parameters, such as transition kernel and reward function, are often time-dependent or even uncertain, but they are approximated as fixed values in an overly simplified setting \citep{mannor2016robust}. Therefore, the output policies of MDPs are often disappointing in practice. 

Robust MDPs address the aforementioned issues of parameter ambiguity, by allowing the unknown values of transition kernels and reward functions to lie in a given ambiguity set \citep{behzadian2021optimizing, chen2019distributionally, clement2021DRMDP, delgado2016real}. Then, robust MDPs seek for policies that maximize the worst-case expected return over all transition kernels and reward functions in the ambiguity sets. By specifying ambiguity sets that contain the unknown transition kernels with high confidence, the optimal policies of robust MDPs are robust to parameter ambiguity \citep{iyengar2005robust}. 

In this paper, we focus on the case where the reward function is ambiguous, which sometimes is referred to as imprecise-reward MDPs \citep{alizadeh2015approximate,regan2010robust,regan2011eliciting,regan2011robust,regan2012regret}. This particular setting is also closely related to imitation~learning, which trains an agent to learn a certain behavior of an expert, while only some demonstrated trajectories of her is available \citep{chen2020bail,ho2016generative,osa2018algorithmic,rashidinejad2021bridging}. When applying inverse reinforcement learning approach to learn the reward function that completely represents the expert's preference \citep{brown2020bayesian,choi2012nonparametric,ng2000algorithms}, the yielded policies, which suffer from reward ambiguity, may perform poorly in practice. 

To handle reward ambiguity, we utilize techniques from distributionally robust optimization (DRO) \citep{derman2020distributional} and distributionally robust chance-constrained program \citep{chen2007robust,postek2018robust}, assuming that the true reward distribution resides in an ambiguity set. This approach does not require the reward function to be precisely specified. Instead, only the descriptions of common distribution information such as support, moments and shape in the ambiguity set are needed, which are often much easier to be obtained/estimated \citep{hanasusanto2015distributionally,hanasusanto2017ambiguous,zymler2013distributionally}. In this paper, we consider a Wasserstein ambiguity set for our distributionally robust models as in \cite{abdullah2019wasserstein, calafiore2006distributionally,xie2021distributionally}.
Unlike phi-divergence ambiguity sets which may contain too extreme member distributions, the closeness between points in the support set is incorporated in Wasserstein sets, thus their member distributions may be more reasonable \citep{gao2022distributionally}; on the other hand, Wasserstein sets are often a better choice than moment-based ambiguity sets when the number of samples is too small to obtain a reliable estimation on moments \citep{yang2020wasserstein}. We choose Wasserstein sets for these reasons, although other types of ambiguity sets such as nested ambiguity sets \citep{xu2010distributionally,xu2012distributionally} and the ambiguity sets based on Prohorov metric \citep{erdougan2006ambiguous} are also considered in literature. For our distributionally robust chance-constrained MDPs, we will furthermore show its equivalence with the nominal counterparts with an adjusted risk level. To the best of our knowledge, this is the first result in MDPs that establishes the mutual transformation between distributional ambiguity and risk.

Our return-risk model (RR) is a risk-averse MDP model that not only takes into account reward ambiguity, but also considers both the average and risk of the return. MDPs that minimize the risk of the return instead of the expected cost are called risk-aware MDPs (also called risk-sensitive or risk-averse MDPs) \citep{ahmadi2021constrained,baauerle2017partially,carpin2016risk, haskell2015convex, huang2017risk}. In risk-aware optimization, the objective function is taken as a risk measure, such as value-at-risk (VaR) \citep{delage2007percentile,delage2010percentile,gilbert2017optimizing}, conditional value-at-risk (CVaR) \citep{bauerle2011markov,chow2017risk,huang2016minimum} and other spectral risk measures \citep{bauerle2021minimizing}, and variants of expected utility   \citep{bernard2022robust, jaimungal2022robust, pflug2007ambiguity}.

\newcontent{
Among these risk measures, VaR and CVaR are arguably the most popular ones and have attracted the attention of many researchers \citep{bauerle2011markov, chow2017risk, delage2007percentile,delage2010percentile,gilbert2017optimizing, huang2016minimum}. By using CVaR, one aims to give a precise depiction of the extreme tail of the distribution (of the uncertain rewards), while VaR does not reflect the extreme scenerios exceeding VaR. It is well-known that CVaR is a coherent risk measure, which can be efficiently optimized by convex optimization tools \citep{chen2021sharing}; in contrast, VaR is a more challenging risk measure because it is
not a coherent one.

One remarkable advantage of VaR is its stability of estimation (especially under fat-tailed reward distribution \citep{sarykalin2008value}), which is particularly important under data-driven settings where the number of samples are limited and decision makers evaluate models based on their out-of-sample performances \citep{bertsimas2006robust, van2022data, zheng2016big}. To demonstrate, we provide an example where we consider a one-step MDP with only 1 state $s$ and 2 actions $a_1$ and $a_2$ \citep{sutton2018reinforcement}. In this one-step MDP, the decision maker only makes one decision in each episode, and she aims to maximize her VaR/CVaR of rewards for these episodes. We consider uncertain rewards $\tilde{r}_{s,a_1}\sim \mathbb{P}_{t\text{-}{\rm dist}}$ and $\tilde{r}_{s,a_2}=\tilde{r}_{s,a_1}+\rho \vert s\vert$ where $\mathbb{P}_{t\text{-}{\rm dist}}$ is a Student's $t$-distribution and we vary its degree of freedom $\delta\in\{2,3,4\}$. We set the shift ratios $\rho=\{0.05i\}_{i\in[5]}$, and for testing the estimation accuracy w.r.t. VaR (resp., CVaR) (where we choose the risk threshold $10\%$), we set the shift quantity $s$ as $\mathbb{P}_{t\text{-}{\rm dist}}\text{-}{\rm VaR}_{0.1}[\tilde{r}_{s,a_1}]$ (resp., $\mathbb{P}_{t\text{-}{\rm dist}}\text{-}{\rm CVaR}_{0.1}[\tilde{r}_{s,a_1}]$), where both risk measures can be efficiently calculated (see Appendix~\ref{apd:t distribution} for more details). We evaluate the decision maker's accuracy rate as the proportion of testing samples where she has chosen the action with a higher VaR/CVaR of rewards (\textit{i.e.}, action $a_2$); for each pair of accuracy rate and shift ratio, following \cite{yamai2002comparative}, 1000 random reward samples for each state-action pair are available for the decision maker, and we test her accuracy rate based on 10000 testing samples.

\begin{figure}[t]
\begin{minipage}[t]{1.0\linewidth}
\centering
\includegraphics[width=6.5in]{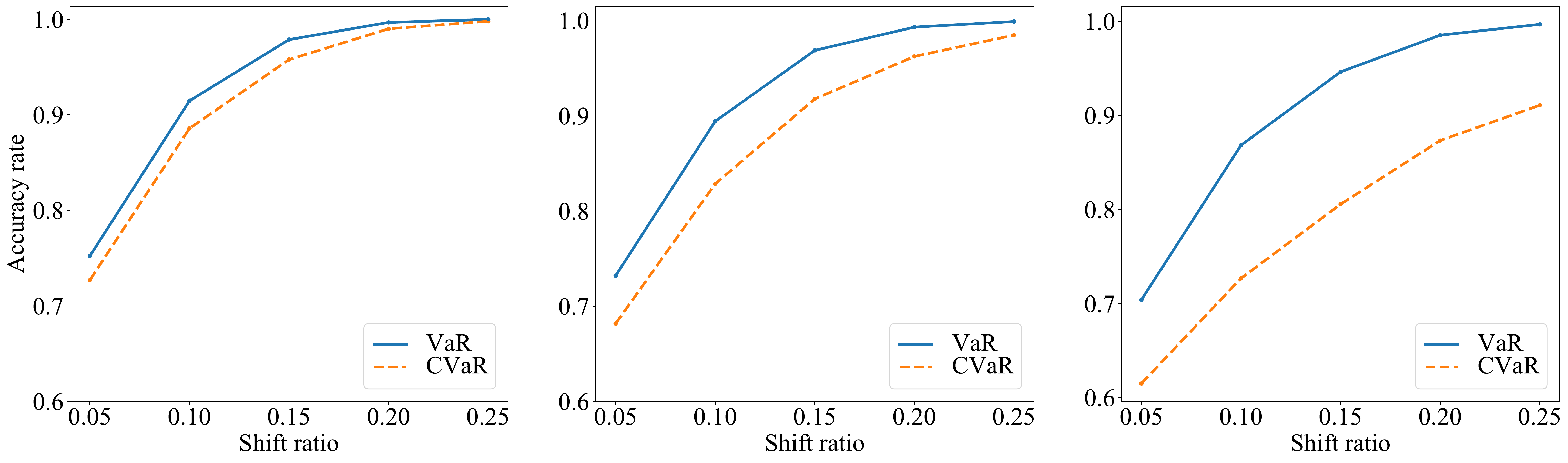}
\end{minipage}
\caption{\textnormal{The accuracy rates of the decision maker choosing the correct action (so that the VaR/CVaR of her rewards is maximized): $\delta=4$ (left), $\delta=3$ (middle) and $\delta=2$ (right).}}
\label{fig:var vs cvar}
\end{figure}

As illustrated in Figure~\ref{fig:var vs cvar}, the accuracy rate increases with the shift ratio $\rho$. As $\delta$ decreases, $\mathbb{F}$ becomes more fat-tailed, and the accuracy rate of VaR is remarkably higher than that of CVaR, which indicates that the statistical inference on VaR would be more accurate than on CVaR. Therefore, VaR may be a more preferable choice when only small sample sets are available.
}


Our return-risk model is motivated by the soft-robust criterion/model, which optimizes a convex combination of the mean and a robust performance in the optimization literature \citep{ben2010soft}. MDPs with soft-robustness are also popular in recent years, where decision makers aim to maximize a weighted average of the mean and percentile performances \citep{brown2020bayesian, lobo2020soft}. Unlike these existing soft-robust MDPs, however, the proposed return-risk model is fundamentally different in two aspects: first, these existing soft-robust models have no consideration for reward ambiguity, while we utilize distributionally robustness to account for reward ambiguity, by which we can hedge against the most adversarial realization of the distribution of rewards (within the ambiguity set), thus our model is more robust to reward uncertainty \citep{chen2019distributionally, xu2010distributionally}; second, we choose VaR as the risk measure which has a direct interpretation to percentile performances, and, as illustrated above, tends to be more advantageous in data-driven optimization.

Our work concentrates on model-based setting, where our proposed models are motivated by the classical (dual formulation of) nominal MDPs \citep{puterman2014markov} and the chance-constrained MDPs \citep{delage2010percentile}. It is worth noting that, beyond model-based setting, there are other inspiring and innovative researches on robust reinforcement learning, such as robust TDC algorithms and robust Q-learning \citep{roy2017reinforcement, wang2021online}, robust policy gradient \citep{wang2022policy}, least squares policy iteration \citep{lagoudakis2003least} and sample complexity analysis \citep{panaganti2022sample}. Note that, though model-free reinforcement learning can be used to learn satisfactory policies for complex environment, the requirement of large amounts of interaction (with environment) may render the learning process slow \citep{kaiser2019model}, while high sample efficiency is one strong advantage of model-based learning \citep{sutton2018reinforcement}. We also note that MDPs with transition kernel ambiguity is another active research line where distributionally robustness is widely employed \citep{clement2021first,shapiro2016rectangular, shapiro2021distributionally, xu2012distributionally}.

We may summarize our contributions as follows (and we also compare our contributions to those of related works in Table~\ref{table:related works} in Appendix~\ref{apd:related works}).



{\it (i)} We show that the distributionally robust model of optimizing expected rewards can be reformulated as a convex conic program, which is equivalent to the nominal MDP with a convex regularization in the objective function.

{\it (ii)} For distributionally robust chance-constrained MDPs (DCC), we show that it can be reformulated as nominal chance-constrained MDPs at adjusted risk levels. This observation bridges the gap between risk and parameter ambiguity.

{\it (iii)} Combining the proposed models in \textit{(i)} and \textit{(ii)}, we propose the return-risk MDP  that maximizes the weighted average of the expectation and VaR of reward (both under distributionally robustness to reward uncertainty), which is flexible and can perform well under the criteria of mean and percentile returns.

\newcontent{
{\it (iv)} When only considering deterministic policies, we show that our return-risk model can also account for risk from uncertain transition kernel, and we derive its equivalent reformulation as a mixed-integer second-order cone program (MISOCP).

}

{\it (v)} To solve the proposed return-risk model, we design a first-order method that is more scalable than the MOSEK solver, thus is faster with large-size problems.

{\it (vi)} In the simulation and empirical experiments, we adopt a data-driven setting, where the decision maker aims at maximizing the expectation and VaR of the random reward. We compare the performances of distributionally robust MDPs (DRMDPs), DCC, RR, \newmodel{robust MDPs (RMDPs) \citep{delage2010percentile} and BROIL \citep{brown2020bayesian}}, and \rr{results show that the third one performs the best under both expectation and different VaR's (with risk thresholds $5\%,\;10\%$ and $15\%$), which showcases its advantages and adjustability to the decision makers' changeable preferences between return and risk}.

The remainder of this paper is organized as follows. We introduce the background in Section~\ref{sec:background}. In Sections~\ref{sec:MDP of expected reward} and~\ref{sec:CC}, we study DRMDPs as well as the DCC model, respectively, and we derive their tractable reformulations. Combining these proposed models, we propose the RR model in Section~\ref{sec: special cases}. \alg{The designed first-order algorithm for the RR model is detailed in Section~\ref{sec:adlpmm}}. We compare the performances of DRMDP, DCC, RR, \newmodel{RMDP and BROIL}, and \alg{demonstrate the advantage of our proposed algorithm} in Section~\ref{sec:numerical study}. Conclusion is drawn in Section~\ref{sec:conclusion}.

\section{Background}\label{sec:background}

We consider an infinite-horizon MDP with a finite state space $\mathcal{S}=\{1,\cdots,S\}$ and a finite action space $\mathcal{A}=\{1,\cdots,A\}$. Let $\bm{P}\in\mathbb{R}^{S \times A \times S}$ be the transition probability kernel such that $p_{s,a,s'}$ is denoted to be the transition probability of transiting to state $s' \in \mathcal{S}$ when action $a\in\mathcal{A}$ is chosen in state $s\in\mathcal{S}$; thus, $\bm{p}_{s,a} \in \mathrm{\Delta}^S$ is the transition probability distribution for every $(s,a) \in \mathcal{S} \times \mathcal{A}$. Given the state-action pair $(s,a)$, an agent will receive an expected reward $r_{s,a} \in \mathbb{R}$. To simplify our notation, we denote the reward function as a vector $\bm{r} = \{r_{s,a}\}_{(s,a)\in \mathcal{S}\times \mathcal{A}}$. 

We seek for the optimal stationary randomized policy $\bm{\pi}=\{\bm{\pi}_s\}_{s\in\mathcal{S}}$ with $\bm{\pi}_s \in \mathrm{\Delta}^A$ for all $s \in \mathcal{S}$, where an action $a\in\mathcal{A}$ will be taken in state $s \in \mathcal{S}$ with probability $\pi_{s,a}$. A nominal MDP that maximizes the expected reward can be formulated \citep{puterman2014markov} as
\begin{equation}\label{prob:nominal MDP}
\ell_{\rm N} = 
\max_{\bm{x} \in \mathcal{X}} \; \bm{r}^\top\bm{x},
\end{equation}
where the feasible set $\mathcal{X}$ is given by
$
\mathcal{X} = \big\{ \bm{x} \in \mathbb{R}_+^{SA} ~\big\vert~ (\bm{E}-\gamma\cdot\bar{\bm{P}})\bm{x} = \bm{p}_0 \big\}.
$
Here the coefficient matrices $\bm{E}={\rm diag}(\bm{e}^\top,\cdots,\bm{e}^\top)\in\mathbb{R}^{S\times SA}$ with $S$ all-ones vectors $\bm{e}\in\mathbb{R}^A$ and $\bar{\bm{P}}=(\bar{\bm{p}}_1,\cdots,\bar{\bm{p}}_S)^\top\in\mathbb{R}^{S\times SA}$ with $\bar{\bm{p}}_s=\{p_{s',a,s}\}_{(s',a)\in\mathcal{S}\times\mathcal{A}}$ for all $s\in\mathcal{S}$.
For each $(s,a) \in \mathcal{S}\times\mathcal{A}$, we denote the $s^{\rm th}$ subvector of $\bm{x}$ as $\bm{x}_s=\{x_i\}_{i \in \{(s-1)A+1,\cdots,sA\}}$; its $a^{\rm th}$ component $x_{s,a}$ can be interpreted as the total discounted probability one occupying state $s$ and choosing action $a$ when applying the policy $\pi^\star_{s,a}=x^\star_{s,a}/(\sum_{a\in\mathcal{A}}x^\star_{s,a})\; \forall (s,a) \in \mathcal{S}\times\mathcal{A}$
\citep{puterman2014markov}\footnotemark\footnotetext{By \cite{puterman2014markov}, any $\bm{x}\in\mathcal{X}$ admits such interpretation, thus we can retrieve our policies of all the proposed models in this paper in this way.}. We have a discount factor $\gamma \in (0,1)$ and the initial distribution $\bm{p}_0\in\mathbb{R}^S_{++}$ of the initial states.  Problem~\eqref{prob:nominal MDP} is a linear program that can be efficiently solved by simplex method and interior-point method \citep{nocedal2006numerical}. One can also compute the optimal policy efficiently by applying value iteration or policy iteration to solve the associated Bellman equation of this problem \citep{bertsekas1995neuro,puterman2014markov}. 

The nominal MDP~\eqref{prob:nominal MDP} does not account for uncertainty in either rewards or transition kernel. To account for reward uncertainty, \cite{delage2010percentile} assume that the random reward vector $\tilde{\bm{r}}$ follows a known Gaussian distribution $\mathbb{P}$ and propose a chance-constrained MDP model as follows:
\begin{equation}\label{prob:chance constrained MDP}
\ell_{\rm CC}(\varepsilon) = \left\{
\begin{array}{c@{\;\;}l@{\;\;}l}
\max & \displaystyle y\\
{\rm s.t.} & \displaystyle \mathbb{P}[\tilde{\bm{r}}^\top\bm{x}\ge y] \ge 1 - \varepsilon \\
& \bm{x} \in \mathcal{X}, \; y \in \mathbb{R}.
\end{array}
\right.
\end{equation}
In fact, the above chance-constrained model maximizes the VaR (at the risk level $1-\varepsilon$) of the reward with respect to the distribution $\mathbb{P}$. Since $\mathbb{P}$ is assumed Gaussian, by theorem~10.4.1 in \cite{prekopa2013stochastic}, one can reformulate problem~\eqref{prob:chance constrained MDP} as a second-order cone program as follows:
\begin{equation}\label{prob:chance constrained MDP reformulation}
\nonumber
\ell_{\rm CC}(\varepsilon) = 
\displaystyle \max_{\bm{x} \in \mathcal{X}} \; \mathbb{E}_{\mathbb{P}}[\tilde{\bm{r}}^\top\bm{x}] - \Vert{\rm F}^{-1}(1-\varepsilon) \bm{\Sigma}^{1/2}\bm{x} \Vert_2,
\end{equation}
where $\mathrm{F}^{-1}(\cdot)$ is the inverse of the cumulative density function of the Gaussian distribution $\mathbb{P}$ and $\bm{\Sigma}$ is the covariance matrix of $\mathbb{P}$. Second-order cone programs allow efficient solutions by state-of-the-art commercial solvers such as CPLEX, Gurobi and MOSEK (see, \textit{e.g.}, \cite{ben2001lectures}). Despite its tractability, the chance-constrained MDP~\eqref{prob:chance constrained MDP} requires the precise underlying reward distribution as input. Moreover, the above reformulation does not hold for generic distribution $\mathbb{P}$.

\section{Distributionally Robust MDPs}\label{sec:MDP of expected reward}

In many real-world situations, the true distribution of the uncertain reward is hard (if not impossible) to obtain. Instead, we may have some firm knowledge, such as moments and shape about it. 
As one of the most efficacious treatments for such situations, the DRO approach models uncertainty as a random variable governed by an unknown probability distribution residing in an ambiguity set. Facing distributional ambiguity, a decision maker seeks for solutions that hedge against the most adversarial distribution from within the ambiguity set. To be specific, in our context, we assume that the true distribution of the uncertain reward resides in a Wasserstein ball of radius $\theta \ge 0$ around some reference distribution $\hat{\mathbb{P}}$: 
\begin{equation}\label{set:Wasserstein ball}
\mathcal{F}(\theta) = \{\mathbb{P} \in \mathcal{P}(\mathbb{R}^{SA}) \mid d_{\mathrm{W}}\big(\mathbb{P}, \hat{\mathbb{P}}\big) \le \theta\}.
\end{equation}
Here $\mathcal{P}(\mathbb{R}^{SA})$ is the set of all probability distributions on $\mathbb{R}^{SA}$, and the Wasserstein distance between two distributions $\mathbb{P}_1$ and $\mathbb{P}_2$, equipped with a general norm $\|\cdot\|$ in $\mathbb{R}^{SA}$, is given by
$
d_{\mathrm{W}}\left(\mathbb{P}_{1}, \mathbb{P}_{2}\right)=\inf _{\mathbb{P} \in \mathcal{Q}\left(\mathbb{P}_{1}, \mathbb{P}_{2}\right)} \mathbb{E}_{\mathbb{P}}[\Vert\tilde{\bm{r}}_{1}-\tilde{\bm{r}}_{2}\Vert],
$
where $\mathcal{Q}(\mathbb{P}_1,\mathbb{P}_2)$ is the set of all joint distributions with marginal distributions $\mathbb{P}_1$ and $\mathbb{P}_2$ that govern $\tilde{\bm{r}}_1$ and $\tilde{\bm{r}}_2$, respectively. 

The random parameter in the nominal MDP~\eqref{prob:nominal MDP} is the expectation of reward, which in practice, is often estimated by the average of historical samples. However, when the sample size is small, such a sample average is not close to the expectation but rather, is known to be optimistically biased (see, \textit{e.g.}, \cite{smith2006optimizer}). Hence, the nominal MDP~\eqref{prob:nominal MDP} based on samples may yield an unsatisfactory policy that does not perform well out-of-sample. For this reason, a possible alternative is to maximize instead the worst-case expected reward as in the following distributionally robust MDP: 
\begin{equation}\label{prob:pessimistic MDP}
\ell_{\rm DRMDP}(\theta) = 
\max_{\bm{x} \in \mathcal{X}} \; \inf_{\mathbb{P}\in\mathcal{F}(\theta)}\mathbb{E}_{\mathbb{P}}[\tilde{\bm{r}}^\top\bm{x}].
\end{equation}
The following proposition offers an equivalent conic program for~\eqref{prob:pessimistic MDP}.

\begin{proposition}\label{prop:pessimistic expected rewards}
The distributionally robust MDP~\eqref{prob:pessimistic MDP} can be reformulated a conic program
\begin{equation}\label{prob:pessimistic MDP reformulation}
\nonumber
\ell_{\rm DRMDP}(\theta) = 
\max_{\bm{x} \in \mathcal{X}} \;  \mathbb{E}_{\hat{\mathbb{P}}}[\tilde{\bm{r}}^\top\bm{x}]-\theta\cdot \Vert\bm{x}\Vert_*.
\end{equation}
\end{proposition}

It is not hard to observe that the distributionally robust MDPs can be viewed as a convex regularization of the nominal MDP~\eqref{prob:pessimistic MDP} under the reference distribution $\hat{\mathbb{P}}$. In particular, the convex regularizing term in the distributionally robust  MDP is $\theta \|\bm{x}\|_*$, which is sized by the Wasserstein radius $\theta$. Interestingly, we have also found that an (distributionally) optimistic MDP can be reformulated as a reverse conic program with a (concave) regularization term $-\theta\Vert\bm{x}\Vert_*$. We relegate this result to Appendix~\ref{apd:optimistic MDPs}.

We remark that, problem~\eqref{prob:pessimistic MDP} is indeed a special case of the robust optimization problem considered in \cite{jaimungal2022robust}, where we consider the expected utility framework. Compared to the policy gradient methods provided in \cite{jaimungal2022robust} where convergence is not established, we have derived its equivalent reformulation as a tractable conic program which can be efficiently solved by state-of-the-art commercial solvers such as Gurobi, Mosek and CPLEX, and can also be seamlessly incorporated in the tractable reformulation of our proposed return-risk model in Section~\ref{sec: special cases}.

\section{Distributionally Robust Chance-Constrained MDPs}\label{sec:CC}
In this section, we turn from optimizing the expectation of reward to its tailed performance, by exploring chance-constrained MDPs. In particular, we still consider Wasserstein ambiguity sets~\eqref{set:Wasserstein ball} to account for distributional ambiguity, meanwhile specifying the reference distribution $\hat{\mathbb{P}}$ and the norm $\|\cdot\|$ in the definition of the Wasserstein distance.

\begin{figure}[tb]
\begin{minipage}[t]{0.33\linewidth}
\centering
\includegraphics[width=2.2in]{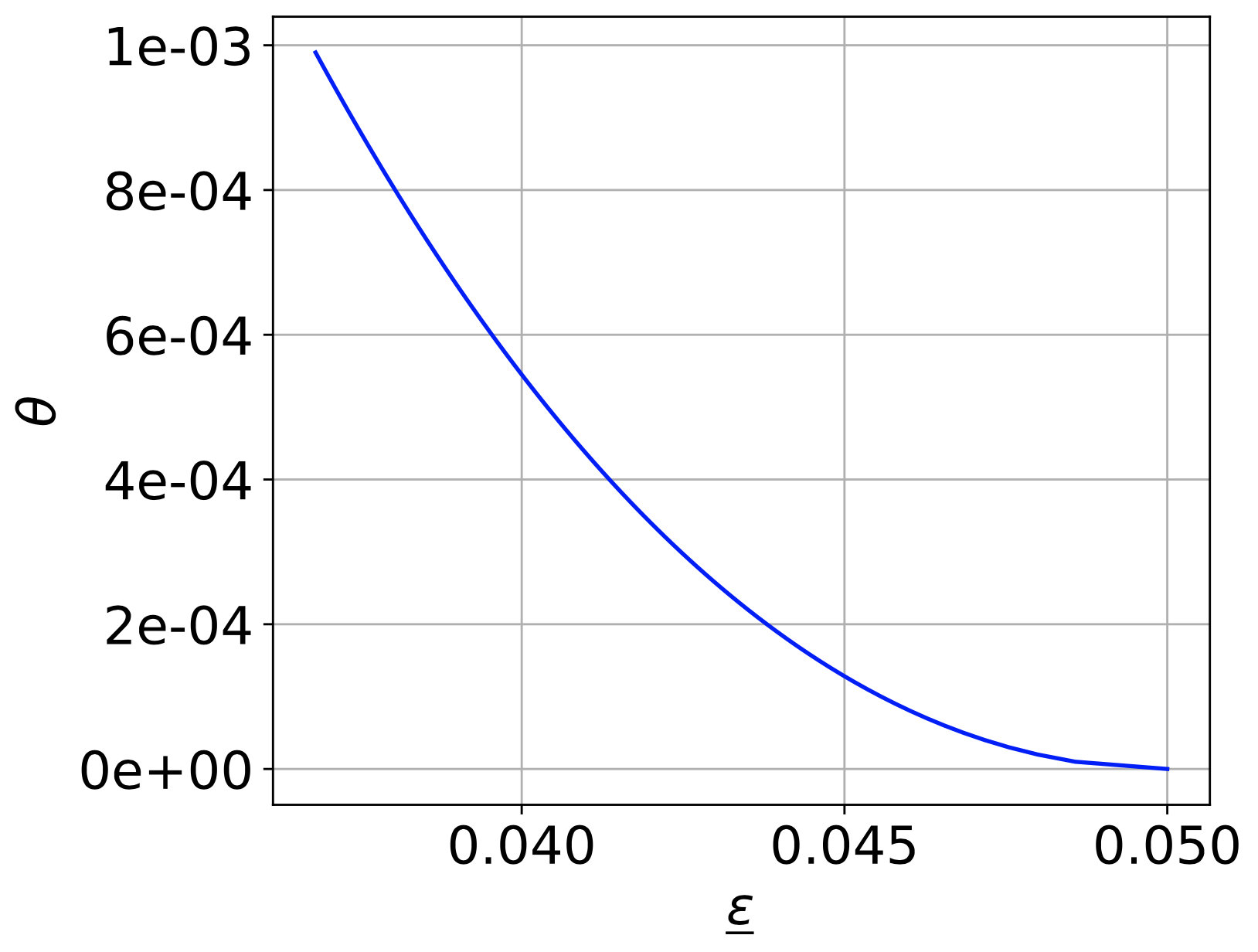}
\end{minipage}%
\begin{minipage}[t]{0.33\linewidth}
\centering
\includegraphics[width=2.2in]{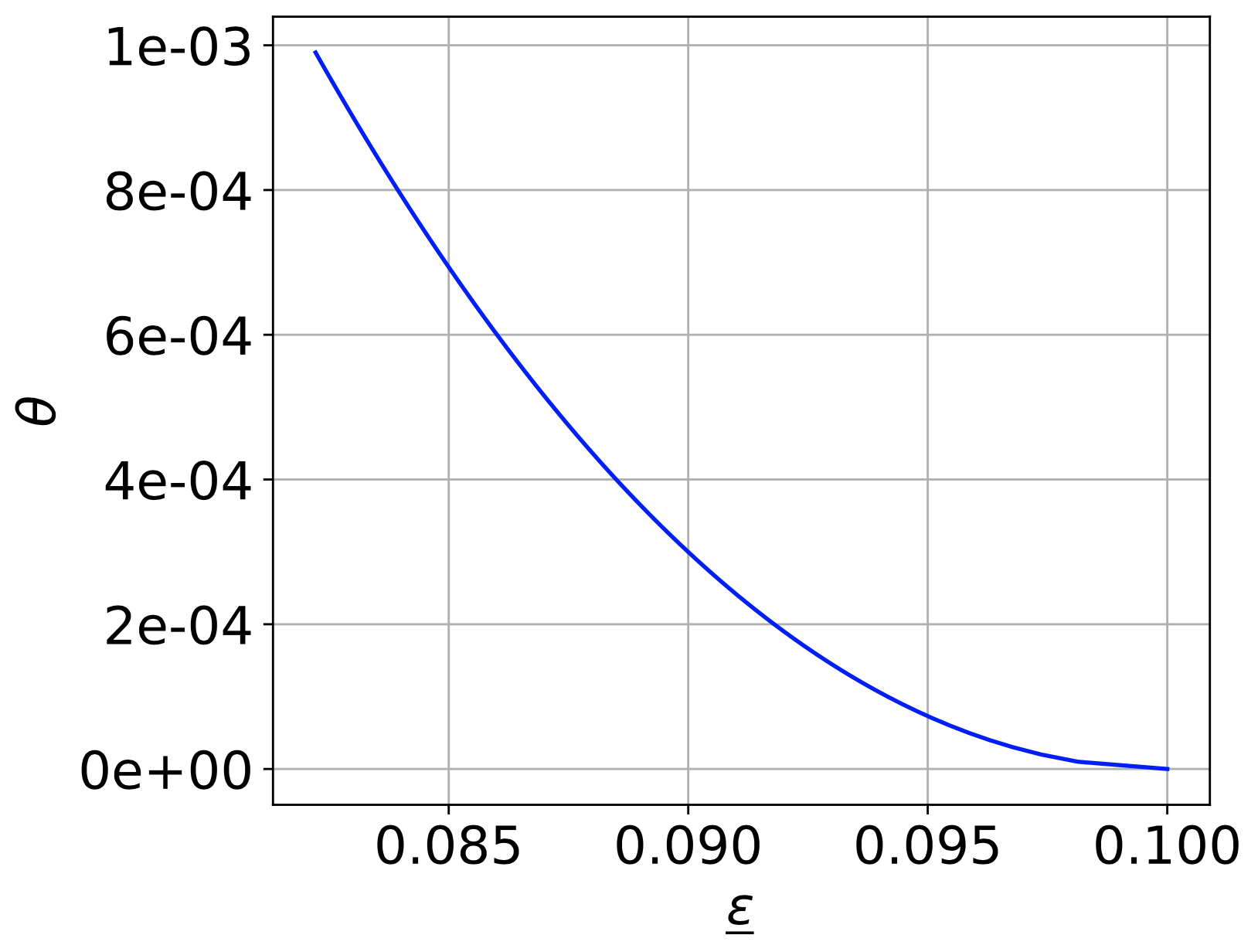}
\end{minipage}
\begin{minipage}[t]{0.33\linewidth}
\centering
\includegraphics[width=2.2in]{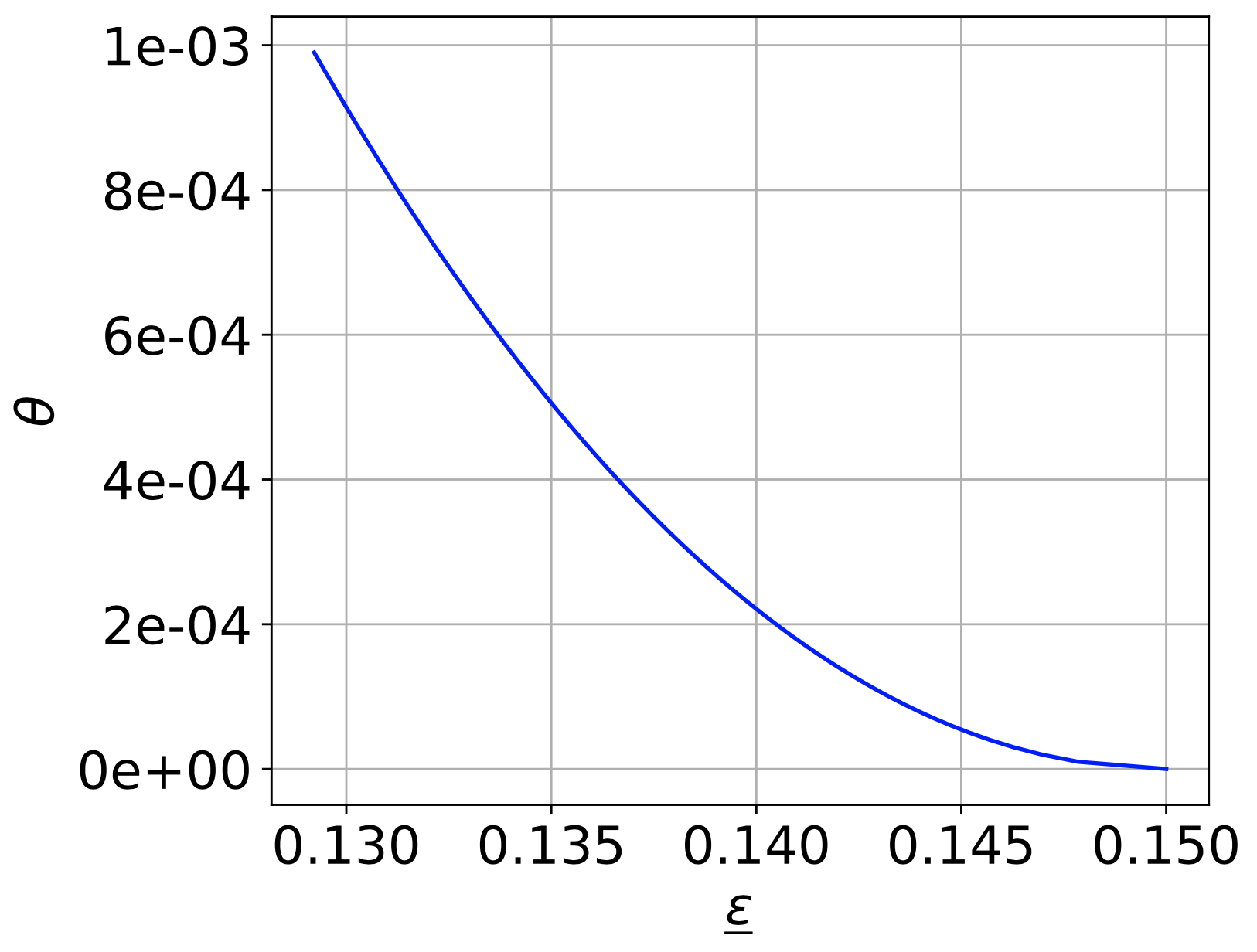}
\end{minipage}
\caption{\textnormal{Values of $\underline{\varepsilon}$ with respect to different $\theta$'s: $\varepsilon = 0.05$ (left), $\varepsilon = 0.1$ (middle), and $\varepsilon = 0.15$ (right).}}
\label{fig:theta and epsilon}
\end{figure}

For the former, we focus on an elliptical reference distribution $\hat{\mathbb{P}}$ = $\mathbb{P}_{(\bm{\mu},\bm{\Sigma},g)}$\footnotemark\footnotetext{Note that results in Section~\ref{sec:MDP of expected reward} hold for a general reference distribution.} throughout this section, whose probability density distribution is given by
$
f(\bm{r})=k \cdot g\left(\frac{1}{2}(\bm{r}-\bm{\mu})^\top \bm{\Sigma}^{-1}(\bm{r}-\bm{\mu})\right),
$
where $k$ is a positive normalization scalar, $\bm{\mu}$ is a mean vector, $\bm{\Sigma}$ is a positive definite matrix and $g$ is a generating function. We emphasize that this assumption on $\hat{\mathbb{P}}$ is mild as this is only the center of the ambiguity set. In particular, our proposed distributionally robust chance-constrained MDPs can account for all types of distributions (as long as they are inside the ambiguity set) and they are not restricted to be all elliptical. As we shall see, such specifications lead to tractable reformulation of our proposed models. Preliminaries on elliptical distributions are relegated to Appendix~\ref{apd:elliptical distributions}.

For the latter, we adopt the Mahalanobis norm associated with the positive definite matrix $\bm{\Sigma}$, captured by $\Vert\bm{x}\Vert_{\bm{\Sigma}} =  \sqrt{\bm{x}^\top\bm{\Sigma}^{-1}\bm{x}}$. Note that the dual norm of a Mahalanobis norm $\Vert\cdot\Vert_{\bm{\Sigma}}$ is another Mahalanobis norm $\Vert\cdot\Vert_{\bm{\Sigma}^{-1}}$ that is defined by the inverse matrix $\bm{\Sigma}^{-1}$.  

In a distributionally robust chance-constrained MDP, we hope that even in the worst-case, with a high confidence the reward is no less than a lower bound, and we aim at maximizing such a lower bound by solving
\begin{equation}\label{prob:pessimistic chance constrained MDP}
\ell_{\rm DCC}(\theta, \varepsilon) = \left\{
\begin{array}{c@{\;\;}l@{\;\;}l}
\max & \displaystyle y\\
{\rm s.t.} & \displaystyle \inf_{\mathbb{P} \in \mathcal{F}(\theta)}\mathbb{P}[\tilde{\bm{r}}^\top\bm{x}\ge y] \ge 1 - \varepsilon \\
& \bm{x} \in \mathcal{X}, \; y \in \mathbb{R}.
\end{array}
\right.
\end{equation}
Quite notably, the worst-case chance constraint in the pessimistic chance-constrained MDP~\eqref{prob:pessimistic chance constrained MDP} is equivalent to a nominal chance constraint in~\eqref{prob:chance constrained MDP} with a higher risky level.

\begin{lemma}\label{lemma:pessimistic chance constraint}
Suppose in the Wasserstein ambiguity set~\eqref{set:Wasserstein ball}, the reference distribution is an elliptical distribution $\hat{\mathbb{P}} = \mathbb{P}_{(\bm{\mu}, \bm{\Sigma}, g)}$ and the Wasserstein distance is equipped with a Mahalanobis norm associated with the positive definite matrix $\bm{\Sigma}$. The distributionally robust chance constraint
\begin{equation}\label{prob:pessimistic chance constraint}
\forall \; \mathbb{P} \in \mathcal{F}(\theta): \mathbb{P}[\tilde{\bm{r}}^\top\bm{x} \ge y] \ge 1-\varepsilon
\end{equation}
is satisfiable if and only if
$
\mathbb{P}_{(\bm{\mu}, \bm{\Sigma}, g)}[\tilde{\bm{r}}^\top\bm{x} \ge y] \ge 1-\underline{\varepsilon},
$
where $\underline{\varepsilon}=1-\mathrm{\Phi}(\bar{\eta}^\star)\le\varepsilon$ with $\bar{\eta}^\star$ that can be computed via bisection method which searches for the smallest $\eta \ge \mathrm{\Phi}^{-1}(1-\varepsilon)$ that satisfies
$
\eta({\rm \Phi}(\eta)-(1-\varepsilon))-\int^{\eta^{2} / 2}_{\left({\rm \Phi}^{-1}(1-\varepsilon)\right)^{2} / 2} k g(z) \mathrm{d} z \ge \theta.
$
\end{lemma}

Equipped with Lemma~\ref{lemma:pessimistic chance constraint}, it then turns out that the distributionally robust chance-constrained MDP~\eqref{prob:pessimistic chance constrained MDP} is equivalent to a nominal chance-constrained MDP~\eqref{prob:chance constrained MDP} at a higher risky level. Consequently, the distributionally robust chance-constrained MDP \eqref{prob:pessimistic chance constrained MDP} can be reformulated into a conic program, or more precisely, a second-order cone program owing to our choice of the Mahalanobis norm. 

\begin{proposition}\label{prop:pessimistic chance constraint socp}
Suppose in the Wasserstein ambiguity set~\eqref{set:Wasserstein ball}, the reference distribution is an elliptical distribution $\hat{\mathbb{P}} = \mathbb{P}_{(\bm{\mu}, \bm{\Sigma}, g)}$ and the Wasserstein distance is equipped with a Mahalanobis norm associated with the positive definite matrix $\bm{\Sigma}$. If the risk threshold satisfies $\varepsilon < 0.5$, then the distributionally robust chance-constrained MDP~\eqref{prob:pessimistic chance constrained MDP} is equivalent to the second-order cone program
$$
\displaystyle \ell_{\rm DCC}(\theta,\varepsilon) = \max_{\bm{x} \in \mathcal{X}} \; \displaystyle \bm{\mu}^\top\bm{x} - \Vert{\rm \Phi}^{-1}(1-\underline{\varepsilon}) \bm{\Sigma}^{1/2}\bm{x} \Vert_2,  
$$
where $\underline{\varepsilon}=1-\mathrm{\Phi}\left(\bar{\eta}^{\star}\right) \le \varepsilon$ with $\bar{\eta}^\star$ being the smallest $\eta \ge \mathrm{\Phi}^{-1}(1-\varepsilon)$ that satisfies
$
\eta({\rm \Phi}(\eta)-(1-\varepsilon))-\int^{\eta^{2} / 2}_{\left({\rm \Phi}^{-1}(1-\varepsilon)\right)^{2} / 2} k g(z) \mathrm{d} z \ge \theta.
$
\end{proposition}

Similar to the distributionally robust MDPs in Section~\ref{sec:MDP of expected reward}, the distributionally robust chance-constrained MDPs also admit an optimistic counterpart, which is equivalent to the nominal chance-constrained MDPs with a larger risk threshold. We relegate this result to Appendix~\ref{apd:optimistic CC}.

To conclude this section, we present in Figure~\ref{fig:theta and epsilon} the relations between $\varepsilon$ and $\underline{\varepsilon}$. Indeed, for any fixed $\varepsilon$, there is a one-to-one correspondence between the risk threshold $\underline{\varepsilon}$ and the Wasserstein radius $\theta$. Following from this fact, for the chance-constrained model in our numerical experiments (Section~\ref{sec:numerical study}), we only calibrate the risk threshold rather than the Wasserstein radius.


\section{Return-Risk MDP}\label{sec: special cases}

For rational decision makers, two types of rewards are their chief concerns: the average and the worst-case rewards. However, the risk-averse models often can not achieve decent average return on which the model put no emphasis \citep{carpin2016risk, delage2010percentile, jiang2018risk}. To take both concerns into considerations,
we leverage the established DRMDPs and DCC model in Sections~\ref{sec:MDP of expected reward} and~\ref{sec:CC} as ingredients and propose the return-risk MDP that maximizes the weighted average of the worst-case expectation and VaR of reward as follows:
\begin{equation}\label{RR}
\begin{array}{r@{\;\;}l}
\ell_{\rm RR}(\alpha, \theta, \varepsilon) =& \displaystyle \max_{\bm{x} \in \mathcal{X}} \; \alpha\inf_{\mathbb{P}\in\mathcal{F}(\theta)}\mathbb{E}_{\mathbb{P}}[\tilde{\bm{r}}^\top\bm{x}] + (1-\alpha)\inf_{\mathbb{P}\in\mathcal{F}'(\theta)}\mathbb{P}\text{-VaR}_{\varepsilon}[\tilde{\bm{r}}^\top\bm{x}].
\end{array}
\end{equation}
Here the Wasserstein ball $\mathcal{F}(\theta)$ is assumed equipped with a general reference distribution and an $L_2$-norm in the definition of the Wasserstein distance, while an elliptical reference distribution $\hat{\mathbb{P}}=\mathbb{P}_{(\bm{\mu},\bm{\Sigma},g)}$ and a Mahalanobis norm associated with the positive definite matrix $\bm{\Sigma}$ are assumed for $\mathcal{F}'(\theta)$. It is not hard to see that the return-risk MDP~\eqref{RR} takes the distributionally robust MDP~\eqref{prob:pessimistic MDP} and the distributionally robust chance-constrained MDP~\eqref{prob:pessimistic chance constrained MDP} in as special cases by varying $\varepsilon$, $\theta$ and $\alpha \in \{0,1\}$. Furthermore, by choosing a fractional $\alpha$, the return-risk model enables one to tailor a balance between risk and return. Proposition~\ref{prop:pessimistic combined MDP} below provides an equivalent second-order cone program for the return-risk MDP~\eqref{RR} under these assumptions.

\begin{proposition}\label{prop:pessimistic combined MDP}
Suppose in~\eqref{RR} the Wasserstein ball $\mathcal{F}(\theta)$ (resp., $\mathcal{F}'(\theta)$) is equipped with a general distribution (resp., an elliptical reference distribution $\hat{\mathbb{P}}=\mathbb{P}_{(\bm{\mu},\bm{\Sigma},g)}$) and the norms in the definitions of the Wasserstein distances of $\mathcal{F}(\theta)$ and $\mathcal{F}'(\theta)$ are an $L_2$-norm and the Mahalanobis norm associated with $\bm{\Sigma}\succ 0$, respectively. Assume that the risk threshold satisfies $\varepsilon < 0.5$, then the return-risk MDP~\eqref{RR} is equivalent to a second-order cone program
\begin{equation}\label{prob:RR reformulation}
\begin{array}{r@{\;\;}l}
\displaystyle \ell_{\rm RR}(\alpha, \theta, \varepsilon) =& \displaystyle\max_{\bm{x} \in \mathcal{X}} \; \bm{\mu}^\top\bm{x}  -\alpha\theta\cdot\Vert\bm{x}\Vert_2  - (1-\alpha)\cdot\Vert{\rm \Phi}^{-1}(1-\underline{\varepsilon}) \bm{\Sigma}^{1/2}\bm{x} \Vert_2,
\end{array}
\end{equation}
where $\underline{\varepsilon}=1-\mathrm{\Phi}\left(\bar{\eta}^{\star}\right) \le \varepsilon$ with $\bar{\eta}^\star$ being the smallest $\eta \ge \mathrm{\Phi}^{-1}(1-\varepsilon)$ that satisfies 
$
\eta({\rm \Phi}(\eta)-(1-\varepsilon))-\int^{\eta^{2} / 2}_{\left({\rm \Phi}^{-1}(1-\varepsilon)\right)^{2} / 2} k g(z) \mathrm{d} z \ge \theta,
$
and it could be computed via bisection method.
\end{proposition}

\newcontent{
\subsection{Risk-Awareness for Uncertain Transition Kernel}

By adopting the static soft-robust framework in \cite{lobo2020soft}, one can indeed also account for the uncertainty in transition kernel in our return-risk model. As in \cite{lobo2020soft}, suppose we have finite samples of transition kernel $\{\hat{\bm{P}}^{i}\}_{i\in[N]}$ with weights $\bm{w}\in\Delta^N:=\{\bm{w}\in\mathbb{R}^N_+\;\vert\;\bm{e}^\top\bm{w}=1\}$ that are generated by MCMC (see, \textit{e.g.}, \cite{kruschke2010bayesian}). Our proposed model is then as follows:
\begin{equation}\label{prob:srr}
\max_{\bm{\pi}\in(\Delta^A)^S}\psi\cdot\mathbb{E}_{\hat{\mathbb{P}}}[g(\bm{\pi}, \bmt{P})] + (1-\psi)\cdot\hat{\mathbb{P}}\text{-}{\rm CVaR}_{\iota}[g(\bm{\pi}, \bmt{P})].
\end{equation}
Here the objective function in \eqref{prob:srr} is again soft-robust against the uncertainty (in transition kernel), with the weight $\psi\in[0,1]$ as the controller for the robustness and $\iota\in[0,1]$ is the risk threshold (w.r.t. the uncertain transition kernel). The weighted empirical distribution $\hat{\mathbb{P}}[\bmt{P}=\hat{\bm{P}}^i]=w_i\;\forall i\in[N]$ and the function 
\begin{equation*}
\begin{array}{rcll}
g(\bm{\pi}, \bm{P}) = &
\max &\bm{\mu}^\top\bm{x}  -\alpha\theta\cdot\Vert\bm{x}\Vert_2 - (1-\alpha)\cdot\Vert{\rm \Phi}^{-1}(1-\underline{\varepsilon}) \bm{\Sigma}^{1/2}\bm{x} \Vert_2\\
&{\rm s.t.}&\displaystyle x_{s,a} = \pi_{s,a}\cdot\sum_{a'\in\mathcal{A}}x_{s,a'}&\forall (s,a)\in\mathcal{S}\times\mathcal{A}\\
&&(\bm{E}-\gamma\cdot\bar{\bm{P}})\bm{x} = \bm{p}_0\\
&&\bm{x}\in\mathbb{R}^{SA}_+
\end{array}
\end{equation*}
represents the optimal value of the return-risk model with the additional constraint that the optimal policy should be the input $\bm{\pi}\in(\Delta^A)^S$ and with $\bar{\bm{P}}$ as the coefficient matrix corresponding to the input transition kernel $\bm{P}$. 

Quite notably, when focusing on deterministic policies, one can reformulate \eqref{prob:srr} as an MISOCP.
\begin{figure*}%
\centering
\begin{equation*}
\boxed{\begin{array}{cll}
\max&\displaystyle (1-\psi)(\eta-\frac{1}{1-\iota}\sum_{i\in[N]}y_i) + \psi 
\cdot\sum_{i\in[N]}(\bm{\mu}^\top\bm{x}^i-\alpha\theta\cdot\Vert\bm{x}^i\Vert_2-(1-\alpha)\Vert{\rm \Phi}^{-1}(1-\underline{\varepsilon}) \bm{\Sigma}^{1/2}\bm{x}^i \Vert_2)\\
{\rm s.t.}&
y_i - w_i \eta \geq \alpha\theta\cdot\Vert\bm{x}^i\Vert_2 + (1-\alpha)\cdot
\Vert{\rm \Phi}^{-1}(1-\underline{\varepsilon}) \bm{\Sigma}^{1/2}\bm{x}^i \Vert_2-\bm{\mu}^\top\bm{x}^i
&\forall i \in[N]\\
&(\bm{E}-\gamma\cdot\bar{\bm{P}}^{i})\bm{x}^i=w_i\cdot\bm{p}_0&\forall i \in[N] \\
&\bm{x}^i\leq \frac{w_i}{1-\gamma}\bm{\pi}&\forall i\in[N] \vspace{1mm}\\
&\displaystyle x^i_{s,a}\geq\frac{w_i}{1-\gamma}(\pi_{s,a}-1)+\sum_{a'\in\mathcal{A}}x^{i}_{s,a'}&\forall (i,s,a)\in \mathcal{N}\times\mathcal{S}\times\mathcal{A}\\
&\bm{\pi}\in(\Delta^A)^S\cap\{0,1\}^{SA},\eta\in\mathbb{R},\bm{x}^i\in\mathbb{R}^{SA}_+,\bm{y}\in\mathbb{R}^N_+&\forall i\in[N].
\end{array}}
\end{equation*}
\caption{\textnormal{Reformulation of \eqref{prob:srr} as an MISOCP.}}
\label{fig:misocp}
\end{figure*}

\begin{proposition}\label{prop:misocp for srr}
If $\bm{\pi}$ is restricted to be a deterministic policy (\textit{i.e.}, $\bm{\pi}\in(\Delta^A)^S\cap\{0,1\}^{SA}$), problem~\eqref{prob:srr} has an equivalent MISOCP reformulation as in Figure~\ref{fig:misocp}.
\end{proposition}

We remark that, though deterministic policies seem to be restricted compared to the randomized ones, they actually are more favored under some situations; for example, they may be a more suitable choice in some medical domains where randomized policies are unworkable for practical and philosophical reasons \citep{rosen2006defense}. Also, randomized policies may be difficult to be evaluated after they have been deployed and may have poor reproducibility \citep{lobo2020soft}. 
}

\section{First-Order Method}\label{sec:adlpmm}
In this section, we introduce an efficient first-order algorithm to solve the equivalent formulation~\eqref{prob:RR reformulation} of our return-risk model.
Our algorithm is based on an alternating direction linearized proximal method of multipliers (AD-LPMM) algorithm \citep{beck2017first, shefi2014rate}, which is a variant of the alternating direction method of multiplier (ADMM) algorithm and also has a convergence rate of $\mathcal{O}(1/N)$ (here $N$ is the number of iterations) proved by \cite{beck2017first}. The proposed splitting allows efficient update of variables in AD-LPMM (where the solutions are analytical or can be retrieved by an efficient bisection method).

For the primal update of the ADMM algorithm, one needs to solve minimization problems with a quadratic term involved (in its objective function); in AD-LPMM, this quadratic term can be linearized by adding a proximity term to the objective function, which could render the primal update much easier. To implement our AD-LPMM algorithm, first we will introduce auxiliary variables and rewrite~\eqref{prob:RR reformulation} (as a minimization problem) as follows:
\begin{equation}\label{prob:RR ADLPMM form}
\begin{array}{c@{\;\;}l@{\;\;}l}
\min & \alpha\theta\cdot\Vert\bm{x}\Vert_2 + (1-\alpha)\cdot\Vert{\rm \Phi}^{-1}(1-\underline{\varepsilon}) \bm{\Sigma}^{1/2}\bm{y} \Vert_2 -\bm{\mu}^\top\bm{z}\\
{\rm s.t.}&(\bm{E}-\gamma\cdot\bar{\bm{P}})\bm{x}=\bm{p}_0\\
&\bm{x}=\bm{y}\\
&\bm{x}=\bm{z}\\
& \bm{x}\in\mathbb{R}^{SA}, \bm{y}\in\mathbb{R}^{SA}, \bm{z}\in\mathbb{R}_{+}^{SA},
\end{array}
\end{equation}
where, in the spirit of AD-LPMM, we can split the decision variables into two groups and update them separately. The augmented Lagrangian function of~\eqref{prob:RR ADLPMM form} is:
\begin{equation*}
\begin{array}{r@{\;\;}l}
&L(\bm{x},\bm{y},\bm{z};\bm{\lambda}, \bm{\xi}, \bm{\eta}) \\ =&\alpha\theta\cdot\Vert\bm{x}\Vert_2 + (1-\alpha){\rm \Phi}^{-1}(1-\underline{\varepsilon})\cdot\Vert \bm{\Sigma}^{1/2}\bm{y} \Vert_2 -\bm{\mu}^\top\bm{z} +\bm{\lambda}^\top((\bm{E}-\gamma\cdot\bar{\bm{P}})\bm{x}-\bm{p}_0)+\bm{\xi}^\top(\bm{x}-\bm{y})\\
&+\bm{\eta}^\top(\bm{x}-\bm{z})+\frac{c}{2}\cdot
\left\Vert
\begin{matrix}
(\bm{E}-\gamma\cdot\bar{\bm{P}})\bm{x}-\bm{p}_0\\
\bm{x}-\bm{y}\\
\bm{x}-\bm{z}
\end{matrix}\right\Vert_2^2.
\end{array}
\end{equation*}
Based on our splitting method, we will update the two groups of variables $(\bm{y},\bm{z})$ and $\bm{x}$ separately. For the update of $(\bm{y},\bm{z})$, we define two primal update operators
\begin{equation*}
\begin{array}{r@{\;\;}l}
\mathfrak{P}_{\bm{y}}(\bm{x},\bm{\xi};c)=\displaystyle \argmin_{\bm{y}}\;(1-\alpha){\rm \Phi}^{-1}(1-\underline{\varepsilon})\cdot\Vert\bm{\Sigma}^{1/2}\bm{y}\Vert_2  -\bm{\xi}^\top\bm{y} + \frac{c}{2}\cdot\Vert\bm{x}-\bm{y}\Vert^2_2
\end{array}
\end{equation*}
and
$
\mathfrak{P}_{\bm{z}}(\bm{x},\bm{\eta};c)=\displaystyle \argmin_{\bm{z}\geq\bm{0}} -\bm{z}^\top(\bm{\mu}+\bm{\eta})+\frac{c}{2}\cdot\Vert\bm{x}-\bm{z}\Vert^2_2;
$
while for the update of $\bm{x}$ (\textit{i.e.}, the second group of variables), we define
\begin{equation*}
\begin{array}{r@{\;\;}l}
\mathfrak{P}_{\bm{x}}(\bm{y},\bm{z},\bm{\lambda},\bm{\xi},\bm{\eta};c,\nu,\hat{\bm{x}}) 
=&\displaystyle\argmin_{\bm{x}}\;\alpha\theta\cdot\Vert\bm{x}\Vert_2+\bm{x}^\top((\bm{E}-\gamma\cdot\bar{\bm{P}})^\top\bm{\lambda}+\bm{\xi}+\bm{\eta})\\
&+\frac{c}{2}\cdot
\left\Vert
\begin{matrix}
(\bm{E}-\gamma\cdot\bar{\bm{P}})\bm{x}-\bm{p}_0\\
\bm{x}-\bm{y}\\
\bm{x}-\bm{z}
\end{matrix}\right\Vert_2^2+\frac{1}{2}\cdot
\ell^2_{\bm{Q}(c,\nu)}(\bm{x}-\hat{\bm{x}}),
\end{array}
\end{equation*}
where
$
\bm{Q}(c, \nu) = c\cdot((\nu-2)\cdot\bm{I}-(\bm{E}-\gamma\cdot\bar{\bm{P}})^\top(\bm{E}-\gamma\cdot\bar{\bm{P}}))
$
and $\ell_{\bm{Q}}(\cdot)$
(equipped with a positive semi-definite matrix $\bm{Q}$) is a weighted vector norm such that $\ell_{\bm{Q}}(\bm{x})=\sqrt{\bm{x}^\top\bm{Q}\bm{x}}$. As we shall see in Section~\ref{subsec:update x}, the update of $\bm{x}$ is fast (where an analytical solution is available) with the proximity term $(1/2)\cdot
\ell^2_{\bm{Q}(c,\nu)}(\bm{x}-\hat{\bm{x}})$ added. Note that when $\bm{Q}(c,\nu)\equiv\bm{0}$, the update in AD-LPMM degenerates to an ADMM's one.

We now introduce our AD-LPMM in Algorithm~\ref{alg:adlpmm}. Basically, the most time-consuming computations lie in the primal update phase, where the updates are carried out by solving a minimization problem with other variables fixed at values after their last updates. As shall be detailed soon, owing to our variable splitting method, the primal updates are also quite fast, where analytical solutions or solutions obtained by bisection are available. Here we choose a stepsize that is increasing in every iteration (with a growth rate $\beta > 0$), which in practice accelerates the convergence.

\begin{algorithm}[t]
\caption{AD-LPMM for Problem~\eqref{prob:RR ADLPMM form} }\label{alg:adlpmm}
\Input{\textnormal{Frobenius norm $\nu=\Vert(\bm{E}-\gamma\cdot\bar{\bm{P}})^\top(\bm{E}-\gamma\cdot\bar{\bm{P}})+2\cdot\bm{I}\Vert_{\rm F}$, initial stepsize $c_0 > 0$, stepsize growth rate $\beta > 0$, desired precision $\delta$,  $\bm{x}^0$, $\bm{y}^0$, $\bm{z}^0$, $\bm{\lambda}^0$, $\bm{\xi}^0$, $\bm{\eta}^0$}}, $k\leftarrow 0$\\
\While{$\left\Vert
\begin{matrix}
(\bm{E}-\gamma\cdot\bar{\bm{P}})\bm{x}^k-\bm{p}_0\\
\bm{x}^k-\bm{y}^k\\
\bm{x}^k-\bm{z}^k
\end{matrix}\right\Vert_{\infty}\geq\delta$}{
\tcp{Primal update}
\textbf{step 1:} $\bm{y}^{k+1} \leftarrow \mathfrak{P}_{\bm{y}}(\bm{x}^k,\bm{\xi}^k;c_k);$

\textbf{step 2:} $\bm{z}^{k+1} \leftarrow \mathfrak{P}_{\bm{z}}(\bm{x}^k,\bm{\eta}^k;c_k)$;

\textbf{step 3:} $\bm{x}^{k+1} \leftarrow \mathfrak{P}_{\bm{x}}(\bm{y}^{k+1},\bm{z}^{k+1},\bm{\lambda}^k,\bm{\xi}^k,\bm{\eta}^k;c_k,\nu,\bm{x}^k)$;

\tcp{Dual update}
\textbf{step 4:} $\bm{\lambda}^{k+1}\leftarrow\bm{\lambda}^k+c_k\cdot((\bm{E}-\gamma\cdot\bar{\bm{P}})\bm{x}^{k+1}-\bm{p}_0)$;\\
\textbf{step 5:} $\bm{\xi}^{k+1}\leftarrow\bm{\xi}^k+c_k\cdot(\bm{x}^{k+1}-\bm{y}^{k+1})$;\\
\textbf{step 6:} $\bm{\eta}^{k+1}\leftarrow\bm{\eta}^k+c_k\cdot(\bm{x}^{k+1}-\bm{z}^{k+1})$;\\
\tcp{Increase stepsize}
\textbf{step 7:} $c_{k+1} \leftarrow c_k + \beta c_0$;\\
\textbf{step 8:} $k \leftarrow k+1$;\\
}
\Output{\textnormal{Solution $\bm{x}^k$}}
\end{algorithm}

\subsection{Subproblem in Step 1: Proximal Mapping and Projection}
To solve $\mathfrak{P}_{\bm{y}}(\bm{x},\bm{\xi};c)$, first we would utilize the technique of proximal mapping and establish the following equivalences:
\begin{equation}\label{prob:y prox}
\begin{array}{r@{\;\;}l}
\mathfrak{P}_{\bm{y}}(\bm{x},\bm{\xi};c)=&{\rm Prox}_{\frac{(1-\alpha){\rm \Phi}^{-1}(1-\underline{\varepsilon})}{c}\cdot\Vert\cdot\Vert_{\bm{\Sigma}}}(\bm{x}+\frac{1}{c}\cdot\bm{\xi})\vspace{1mm}\\
=&\bm{x}+\frac{1}{c}\cdot\bm{\xi}-\frac{(1-\alpha){\rm \Phi}^{-1}(1-\underline{\varepsilon})}{c}\cdot{\rm Proj}_{\bm{B}_{\ell_{\bm{\Sigma}^{-1}}(\cdot)}}\left(\frac{1}{(1-\alpha){\rm \Phi}^{-1}(1-\underline{\varepsilon})}\cdot(c\cdot\bm{x}+\bm{\xi})\right),
\end{array}
\end{equation}
where 
$
{\rm Prox}_{f(\cdot)}(\bm{x})=\argmin_{\bm{v}}f(\bm{v})+\frac{1}{2}\cdot\Vert\bm{v}-\bm{x}\Vert^2_2
$
is the proximal mapping operator and
\begin{equation}\label{prob:projection}
{\rm Proj}_{\bm{B}_{\ell_{\bm{\Sigma}}(\cdot)}}(\bm{x})=\displaystyle\argmin_{\bm{v}:\ell_{\bm{\Sigma}}(\bm{v})\leq 1} \frac{1}{2}\cdot\Vert\bm{v}-\bm{x}\Vert_2^2
\end{equation}
is the operator of projection on the unit ball
$
\bm{B}_{\ell_{\bm{\Sigma}}(\cdot)}=\{\bm{x}\in\mathbb{R}^{SA}\;\vert\;{\ell_{\bm{\Sigma}}(\bm{x})}\leq 1\}.
$
Here, the first equality in~\eqref{prob:y prox} holds by the definition of the proximal mapping operator, and the second equality follows from,\textit{e.g.}, example~$6.4.7$ in \cite{beck2017first}. Indeed, problem \eqref{prob:projection} allows an efficient solution obtained by a bisection method to locate its optimal dual solution $\lambda^\star\geq 0$ (after which the optimal primal solution can be retrieved immediately),
where the upper bound of the bisection is provided in Lemma~\ref{lemma:bisect ub} relegated to Appendix~\ref{apd:proof of adlpmm}. The time complexity of the solution process~\eqref{prob:y prox}, as well as the pseudocode for the bisection method, are provided in the following proposition.
\begin{proposition}\label{prop:projection}
Problem $\mathfrak{P}_{\bm{y}}(\bm{x},\bm{\xi};c)$ can be solved in time $\mathcal{O}(SA\log(1/\delta'))$, where $\delta'$ is the desired precision of the bisection method.
\end{proposition}

\subsection{Subproblem is Step 2: Componentwise Update}
Problem $\mathfrak{P}_{\bm{z}}(\bm{x},\bm{\eta};c)$ can be decomposed into $SA$ single-variable quadratic programming problems, each allowing an analytical solution. We summarize the time complexity and details in the following proposition.
\begin{proposition}\label{prop:z}
Problem $\mathfrak{P}_{\bm{z}}(\bm{x},\bm{\eta};c)$ can be solved in time $\mathcal{O}(SA)$.
\end{proposition}

\subsection{Subproblem in Step 3: Linearization and Proximal Mapping}\label{subsec:update x}
Compared to the update in ADMM, in our AD-LPMM, a proximity term
$(1/2)\cdot\ell^2_{\bm{Q}(c,\nu)}(\bm{x}-\hat{\bm{x}})$ is added to the objective function of the update in step~$3$. By choosing $\bm{Q}(\cdot,\cdot)$ as mentioned in Section~\ref{sec:adlpmm}, we can linearize all the quadratic terms in $\mathfrak{P}_{\bm{x}}(\bm{y},\bm{z},\bm{\lambda},\bm{\xi},\bm{\eta};c,\nu,\hat{\bm{x}})$, thus the solution can be obtained analytically by the technique of proximal mapping (meanwhile assuring the positive semi-definiteness of $\bm{Q}(c_k,\nu)$ in every iteration of Algorithm~\ref{alg:adlpmm}). This solution process, as well as its time complexity, is provided in the following proposition.

\begin{proposition}\label{prop:x}
Problem $\mathfrak{P}_{\bm{x}}(\bm{y},\bm{z},\bm{\lambda},\bm{\xi},\bm{\eta};c,\nu,\hat{\bm{x}})$ can be solved in time $\mathcal{O}(SA)$.
\end{proposition}

\section{Numerical Experiments}\label{sec:numerical study}
In this section, we conduct two numerical experiments to compare the performances of DRMDPs~\eqref{prob:pessimistic MDP}, CC~\eqref{prob:chance constrained MDP}\footnotemark\footnotetext{As we demonstrated in Section \ref{sec:CC}, a DCC is equivalent to a nominal chance-constrained one with an adjusted risk level, thus here we simply choose the latter as the benchmark.}, RR~\eqref{RR}, \newmodel{RMDPs \citep{delage2010percentile} and BROIL \citep{brown2020bayesian} (please see  Appendices~\ref{apd:RMDP} and \ref{apd:BROIL} for more details for the last two models)}. In both experiments, we train our reward functions with different sample sizes (100,200,300,400,500). For each sample size, performance of each model is evaluated for 100 times. \rr{The performance of each model is evaluated by expectation and VaR with risk thresholds $\varepsilon'\in\{5\%, 10\%, 15 \%\}$}. Cross validations are conducted for parameter selection (please see Appendix~\ref{apd:parameter} for details). 

In Section~\ref{sec:simulation}, we conduct a simulation study where MDPs are generated randomly as in \cite{regan2012regret}; In Section~\ref{sec:machine replacement}, we study a machine replacement problem introduced in \cite{delage2010percentile}. 
As implied in our proofs, in this section, the Wasserstein ambiguity set of DRMDPs~\eqref{prob:pessimistic MDP} will be equipped with a general reference distribution and an $L_2$-norm for the Wasserstein distance; as for RR~\eqref{RR}, we use a general reference distribution and an $L_2$-norm in the definition of the Wasserstein distance for the Wasserstein ambiguity set $\mathcal{F}(\theta)$, while for $\mathcal{F'}(\theta)$, we use an elliptical reference distribution $\hat{\mathbb{P}}=\mathbb{P}_{(\bm{\mu},\bm{\Sigma},g)}$ and the Mahalanobis norm associated with the positive definite matrix $\bm{\Sigma}$ for the Wasserstein distance. All optimization problems are solved by MOSEK on a $2.3$GHz processor with $32$GB memory.
\begin{figure}[tb]
\begin{minipage}[t]{0.5\linewidth}
\centering
\includegraphics[width=3in]{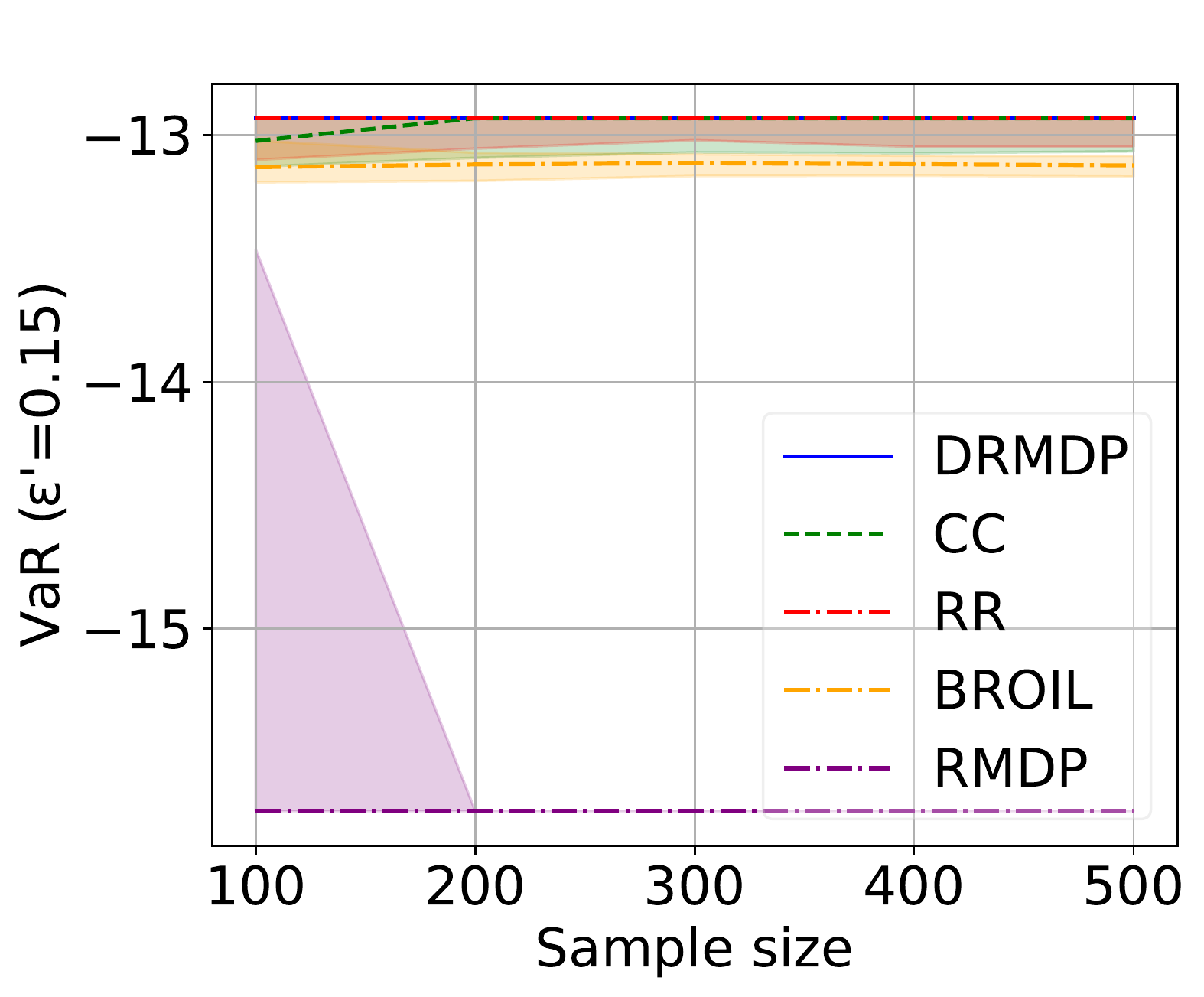}
\end{minipage}%
\begin{minipage}[t]{0.5\linewidth}
\centering
\includegraphics[width=3in]{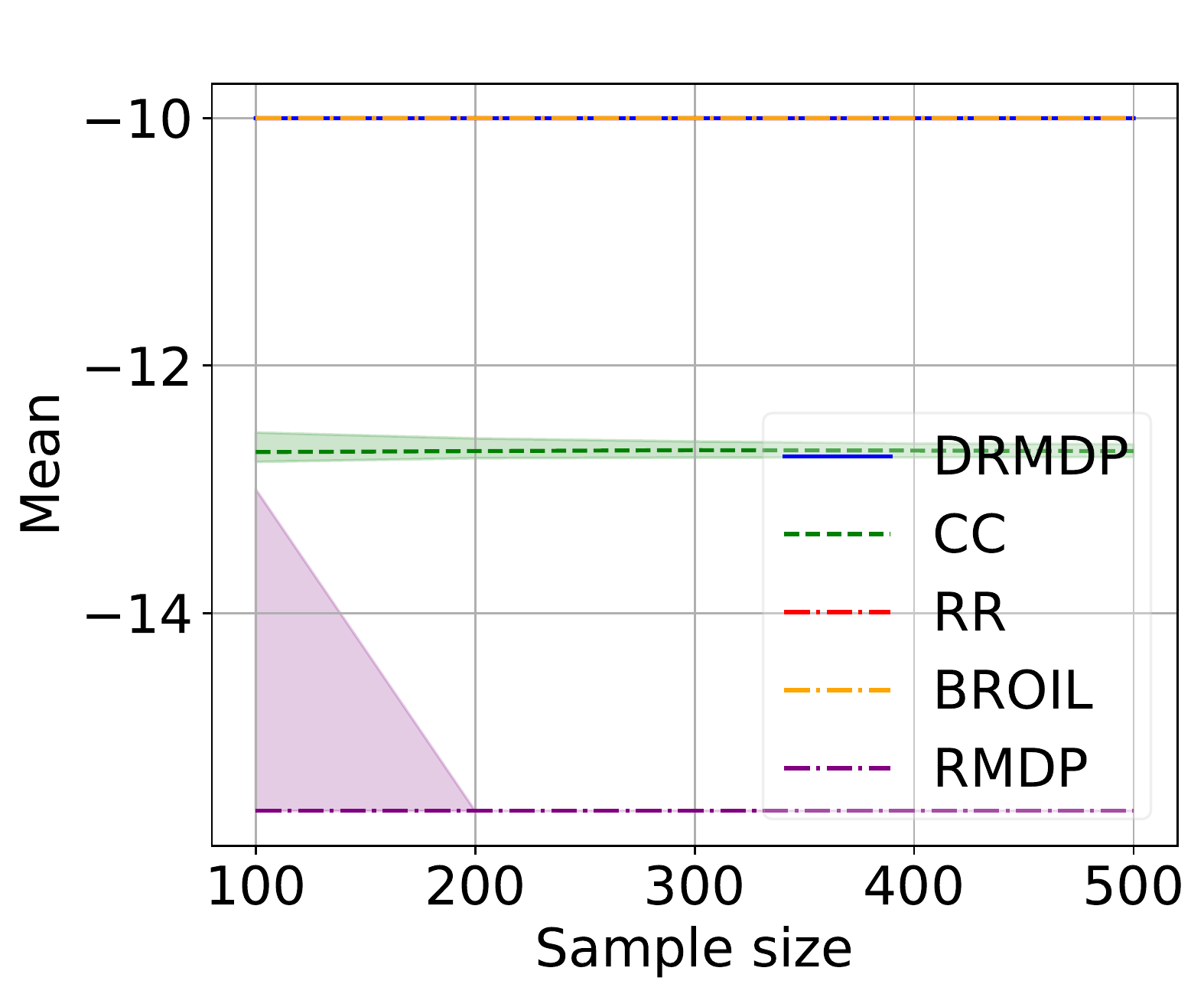}
\end{minipage}
\caption{\textnormal{Empirical study. Models DRMDP~\eqref{prob:pessimistic MDP}, CC~\eqref{prob:chance constrained MDP}, RR~\eqref{RR}, RMDP and BROIL evaluated by VaR (risk threshold $\varepsilon'=15\%$) and mean of reward. The upper and lower edges of the shaded areas are respectively the 95\% and 5\% percentiles of the 100 performances, while the solid lines are the medians.}}
\label{fig:empirical 01}
\end{figure}

\subsection{Simulation Study}\label{sec:simulation}
\begin{figure}[tb]
	\begin{minipage}[t]{0.5\linewidth}
		\centering
		\includegraphics[width=3in]{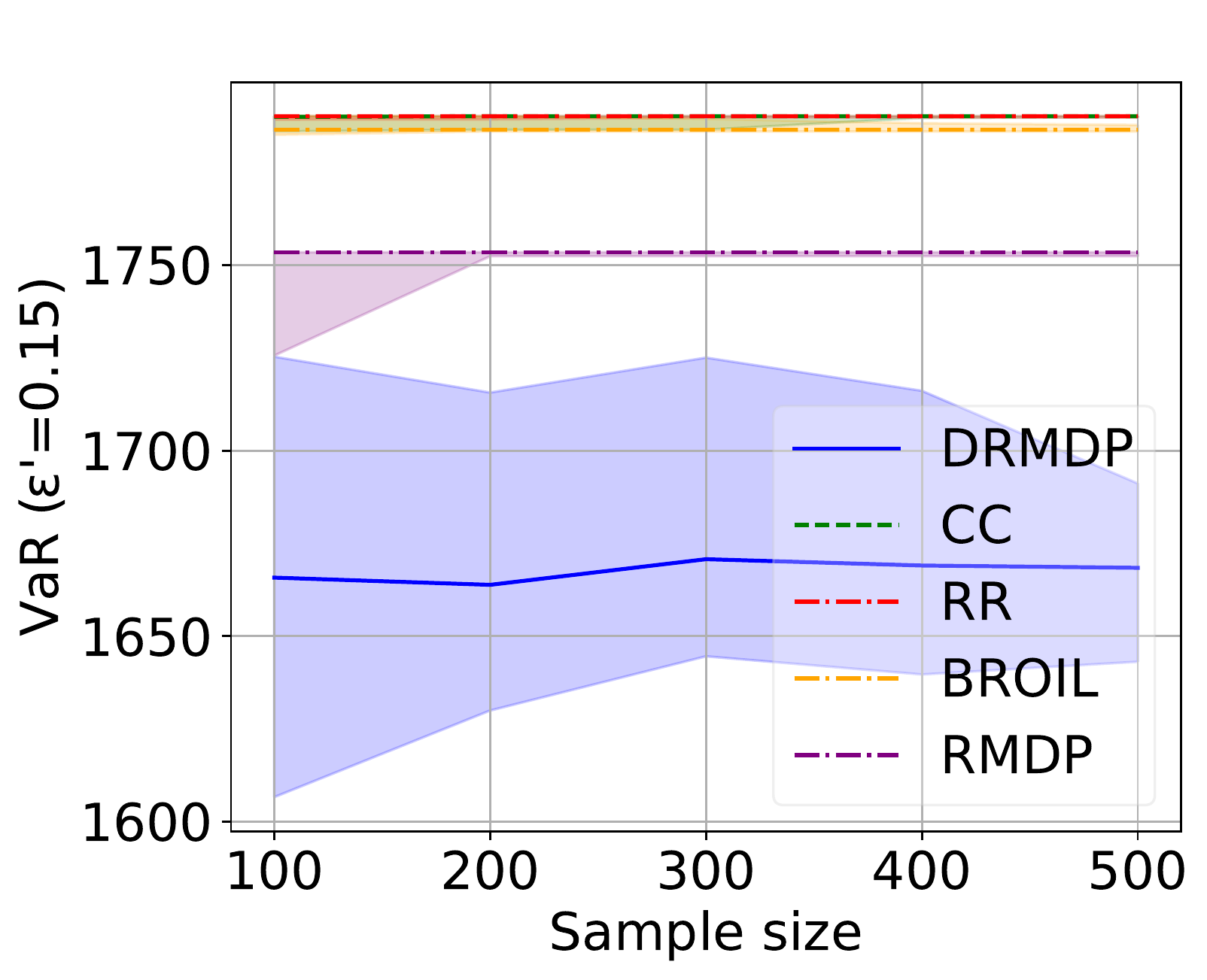}
	\end{minipage}%
	\begin{minipage}[t]{0.5\linewidth}
		\centering
		\includegraphics[width=3in]{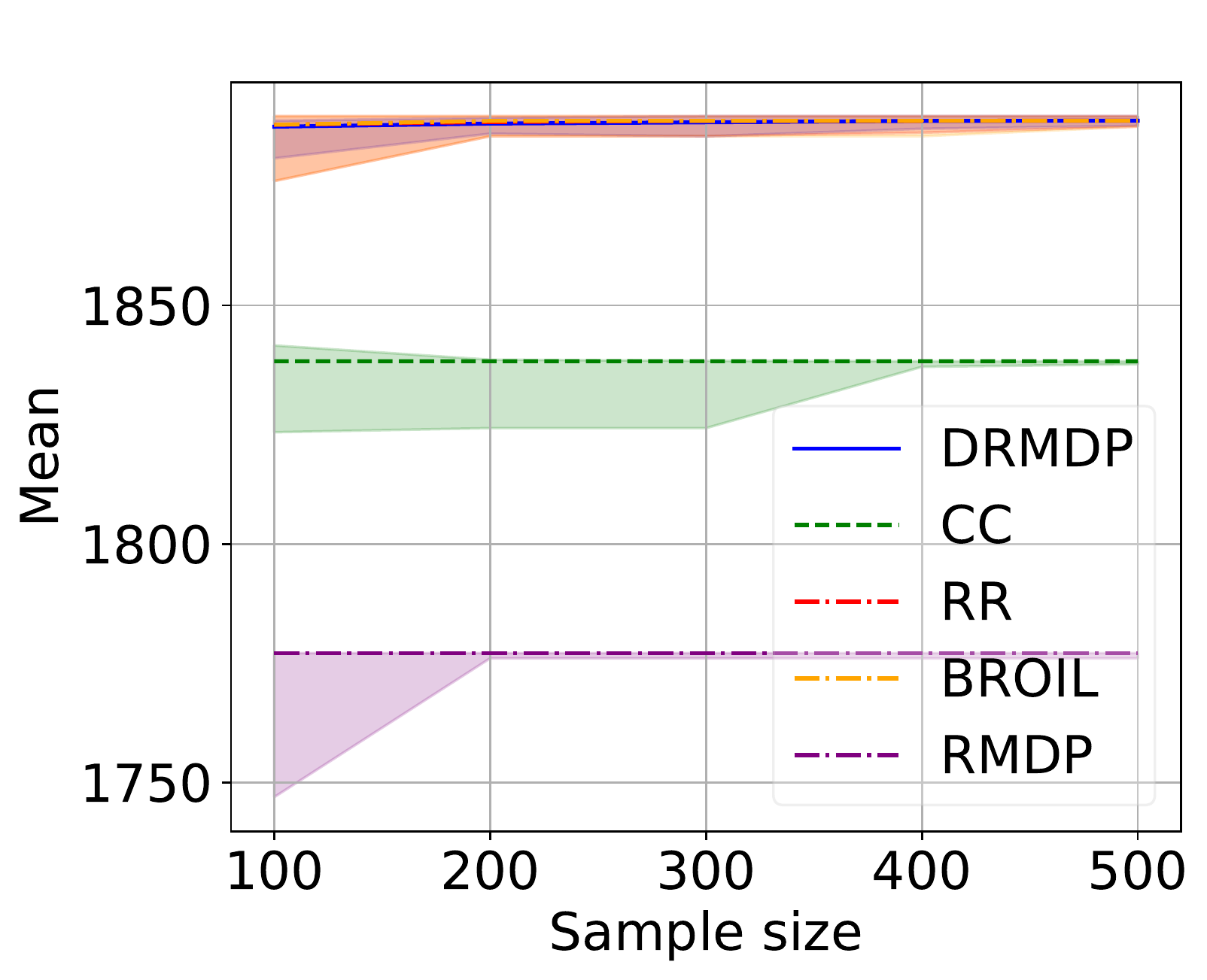}
	\end{minipage}
	\caption{\textnormal{Simulation. Models DRMDP~\eqref{prob:pessimistic MDP}, CC~\eqref{prob:chance constrained MDP}, RR~\eqref{RR}, RMDP and BROIL evaluated by VaR (risk threshold $\varepsilon'=15\%$) and mean of reward. The upper and lower edges of the shaded areas are respectively the 95\% and 5\% percentiles of the 100 performances, while the solid lines are the medians.}}
	\label{fig:simulation 01}
\end{figure}
In this experiment, we follow the experiment setup in \cite{regan2012regret} where the number of reachable next-states and the transition kernel are randomly generated (both of which are known to decision makers). More details of the experiment setting are relegated to Appendix~\ref{apd:simulation}.

\rr{As illustrated in Figures~\ref{fig:simulation 01} and \ref{fig:simulation 05 10} (where the latter for VaR with $\varepsilon'\in\{5\%,10\%\}$ is relegated to Appendix~\ref{apd:simulation results}), when the decision maker aims to optimize her tailed performances, CC is a preferable choice compared to DRMDPs; on the contrary, when pursuing optimizing the average return, DRMDPs perform much better than CC. Observe that the RR model, which includes both DRMDPs and the DCC model as special cases, remains as the best model under all criteria. In particular, one can observe that, RR achieves higher percentile returns than BROIL (that is a model without robustness), which demonstrates the benefits of distributionally robustness and the advantage of the risk measure VaR for percentile performance optimization. As expected, RMDPs end up yielding over-conservative policies; as a result, it performs poorly in most instances under all criteria}.

\subsection{Machine Replacement Problem}\label{sec:machine replacement}
In this experiment, we follow the experiment setup in \cite{delage2010percentile} and consider the case where a factory holds an extensive amount of machines, each of which is subject to the same underlying MDP (more details of the experiment setting can be found in Appendix~\ref{apd:simulation}). Our setting is similar to \cite{delage2010percentile} except for the follows: we use a data-driven setting as described above, and we evaluate our (policies of) models by looking at the various performance measures as in Section~\ref{sec:simulation}.

\rr{We report the overall performances of the five models in Figures~\ref{fig:empirical 01} and \ref{fig:empirical 05 10} (where the latter for VaR with $\varepsilon'\in\{5\%,10\%\}$ is relegated to Appendix~\ref{apd:empirical results}). Similar to the previous experiment,
RR always performs better than or equal to the best model between CC and DRMDPs, and it provides the best performance under all criteria, which again manifest the merit of taking both the expected and worst-case performances into consideration and distributionally robustness.}

\subsection{Computation Times of Different Algorithms}
\begin{table}[h]
\caption{\textnormal{The average of the runtimes of the MOSEK solver and the AD-LPMM algorithm in seconds and the relative gaps ($\%$) to the optimal values computed by MOSEK.}}
\centering
\begin{tabular}{cccrrr}
\toprule
\multirow{2}{*}{\tabincell{c}{S=A}}& \multicolumn{2}{c}{Runtimes} &  \multirow{2}{*}{\tabincell{c}{Relative gaps}} \\\cline{2-3}
& MOSEK & AD-LPMM &  \\ \midrule
40      &   \textbf{0.60}      &   2.79       & $<$ 0.1 $\%$      \\
70      &   5.58      &   \textbf{4.81}       & $<$ 0.1    $\%$    \\
100      &   25.50      &   \textbf{19.98}       & 0.2   $\%$    \\
130      & 93.54     &     \textbf{66.17}      & $<$ 0.1 $\%$  \\
160      &     444.06 &    \textbf{168.34}     &     0.4 $\%$  \\
\bottomrule 
\end{tabular} \label{table:computation times}
\end{table}

In this section, we compare the computation times of our AD-LPMM algorithm with the state-of-the-art solver MOSEK. Table~\ref{table:computation times} reports the runtimes of the the AD-LPMM and MOSEK when solving problem~\eqref{prob:RR reformulation} at different problem sizes. Results indicate that, though our AD-LPMM is slower than the MOSEK solver when problem size is small, it showcases its strong scalability and become much faster than MOSEK with large-size problems (while always maintaining high solution quality), where the advantage is more notable when the problem scales up.

\section{Conclusion}\label{sec:conclusion}
We consider risk-aware MDPs with ambiguous reward functions and propose the return-risk model, which is versatile and can optimize any weighted combination of the average and quantile performances of a policy. This model generalizes and combines the advantage of  distributionally robust MDPs and distributionally robust chance-constrained MDPs, thus is powerful in both average and percentile performances optimization. \newcontent{In particular, risk from uncertain transition kernel can also be captured by the return-risk model when output policies are deterministic.}
Tractable reformulations are provided for all our proposed models, and \alg{we design an AD-LPMM algorithm for the return-risk model, which is well scalable and faster than the MOSEK solver with large-scale problems.} Experimental results showcase the \rr{versatility of the return-risk model}
as well as \alg{the scalability of the algorithm}.


In the future, we believe that it would be important to explore more efficient methods for obtaining solution of RR, where function approximation and policy gradient \citep{sutton2018reinforcement} are possible choices to achieve this.

\bibliographystyle{ormsv080}
\bibliography{References}

\newpage
\begin{appendices}
\section{Proof of Results}

\subsection{Proofs of Results in Section~\ref{sec:MDP of expected reward}}

\noindent \textit{Proof of Proposition~\ref{prop:pessimistic expected rewards}.} $\;$
It is sufficient to rewrite the objective of \eqref{prob:pessimistic MDP} as follows:
\[
\begin{array}{r@{\;\;}l}
\displaystyle \inf_{\mathbb{P}\in\mathcal{F}(\theta)} \mathbb{E}_\mathbb{P}[\tilde{\bm{r}}^\top\bm{x}]
& = \displaystyle  -\sup_{\mathbb{P}\in\mathcal{F}(\theta)} \mathbb{E}_\mathbb{P}[-\tilde{\bm{r}}^\top\bm{x}] \\[3mm]
& = \displaystyle -\min_{\lambda \ge 0} \bigg\{\lambda\theta - \int_{\mathbb{R}^{SA}}\inf_{\bm{\xi}\in\mathbb{R}^{SA}}(\lambda\Vert\bm{\xi}-\bm{r}\Vert+ \bm{\xi}^\top\bm{x}) \;{\rm d}\hat{\mathbb{P}}\bm{r} \bigg\} \\[3mm]
& =  \displaystyle -\min_{\lambda \ge \Vert\bm{x}\Vert_*} \bigg\{\lambda\theta - \int_{\mathbb{R}^{SA}}\bm{r}^\top\bm{x} \; {\rm d}\hat{\mathbb{P}}\bm{r} \bigg\} \\[3mm]
& = \displaystyle \mathbb{E}_{\hat{\mathbb{P}}}[\tilde{\bm{r}}^\top\bm{x}]-\theta\Vert\bm{x}\Vert_*,
\end{array}
\]

where the second identity follows from theorem~1 in \cite{gao2016distributionally} and the third identity follows from strong conic duality 
\[
\inf_{\bm{\xi}\in\mathbb{R}^{K}}(\lambda\Vert\bm{\xi}-\bm{r}\Vert+ \bm{\xi}^\top\bm{x}) = \left\{
\begin{array}{l@{\;\;}l}
\bm{r}^\top\bm{x} &~~\lambda \ge \Vert\bm{x}\Vert_* \\
-\infty &~~ \lambda \in [0,\Vert\bm{x}\Vert_*).
\end{array}
\right.
\]
Substituting the above reexpression then concludes the proof.
\hfill\qed

\subsection{Proofs of Results in Section~\ref{sec:CC}}

\textit{Proof of Lemma~\ref{lemma:pessimistic chance constraint}.} $\;$
Notice that \eqref{prob:pessimistic chance constraint} is equivalent to 
\[
\sup _{\mathbb{P} \in \mathcal{F}(\theta)} \mathbb{P}\left[\tilde{\bm{r}}^\top\bm{x} < y\right] \le \varepsilon \Longleftrightarrow \sup _{\mathbb{P} \in \mathcal{F}(\theta)} \mathbb{P}\left[\tilde{\bm{r}}^\top\bm{x} \le y \right] \le \varepsilon,
\]
where it is equivalent if we replace the strict inequality on the left-hand side with a weak one on the right-hand side; see proposition 3 in \cite{gao2016distributionally}. Exploring the definition of VaR, we note that
\[
\sup _{\mathbb{P} \in \mathcal{F}(\theta)} \mathbb{P}\left[\tilde{\bm{r}}^\top\bm{x} \le y\right] \le \varepsilon \Longleftrightarrow \sup _{\mathbb{P} \in \mathcal{F}(\theta)} \mathbb{P}\text{-VaR}_{1-\varepsilon}\left[y-\tilde{\bm{r}}^\top\bm{x} \right] \le 0.
\]
By corollary~4.9 in \cite{chen2021sharing} and the assumption of Mahalanobis norm, it holds that 
\begin{equation*}
\sup _{\mathbb{P} \in \mathcal{F}(\theta)} \mathbb{P}\text{-VaR}_{1-\varepsilon}\left[y-\tilde{\bm{r}}^\top\bm{x}\right]=\mathbb{P}_{(\bm{\mu}, \mathbf{\Sigma}, g)}\text{-VaR}_{1-\underline{\varepsilon}}\left[y-\tilde{\bm{r}}^\top\bm{x}\right].
\end{equation*}
In other words, the worst-case VaR around the elliptical distribution $\mathbb{P}_{(\bm{\mu}, \mathbf{\Sigma}, g)}$ with the risk threshold $\varepsilon$ is equal to the nominal elliptical VaR with a small risk threshold $\underline{\varepsilon} \le \varepsilon$ (which, would correspond to a higher risk level). We thus obtain
\begin{equation*}
\begin{array}{c@{\;\;}l}
\displaystyle \sup _{\mathbb{P} \in \mathcal{F}(\theta)} \mathbb{P}\text{-VaR}_{1-\varepsilon}\left[y-\tilde{\bm{r}}^\top\bm{x}\right] \le 0 & \Longleftrightarrow \mathbb{P}_{(\bm{\mu}, \mathbf{\Sigma}, g)}\text{-VaR}_{1-\underline{\varepsilon}}\left[y-\tilde{\bm{r}}^\top\bm{x}\right] \le 0 \\
&\displaystyle \Longleftrightarrow \mathbb{P}_{(\bm{\mu}, \mathbf{\Sigma}, g)}\left[\tilde{\bm{r}}^\top\bm{x} \le y \right] \le \underline{\varepsilon} \\[3mm] 
&\displaystyle \Longleftrightarrow \mathbb{P}_{(\bm{\mu}, \mathbf{\Sigma}, g)}\left[\tilde{\bm{r}}^\top\bm{x} \ge y \right] \ge 1-\underline{\varepsilon},
\end{array}
\end{equation*}
where the last equivalence follows from $\mathbb{P}_{(\bm{\mu}, \mathbf{\Sigma}, g)}$ being a continuous distribution. 
\hfill\qed

\textit{Proof of Proposition~\ref{prop:pessimistic chance constraint socp}.} $\;$
By Lemma~$1$, the first constraint in \eqref{prob:pessimistic chance constrained MDP} is the same as 
\begin{equation*}
\mathbb{P}_{(\bm{\mu}, \bm{\Sigma}, g)}\left[\tilde{\bm{r}}^\top\bm{x}\ge y\right] \ge 1 - \underline{\varepsilon},
\end{equation*}
where $\underline{\varepsilon}=1-\mathrm{\Phi}(\bar{\eta}^\star)\le\varepsilon$ and $\bar{\eta}^\star$ is the smallest $\eta \ge \mathrm{\Phi}^{-1}(1-\varepsilon)$ that satisfies
\begin{equation*}
\eta({\rm \Phi}(\eta)-(1-\varepsilon))-\int^{\eta^{2} / 2}_{\left({\rm \Phi}^{-1}(1-\varepsilon)\right)^{2} / 2} k g(z) \mathrm{d} z \ge \theta.
\end{equation*}
The constraint can then be further written as
\[
\begin{array}{r@{\;\;}l}
\mathbb{P}_{(\bm{\mu},\bm{\Sigma},g)}[\tilde{\bm{r}}^\top\bm{x} \ge y] \ge 1-\underline{\varepsilon}
\Longleftrightarrow & {\rm \Phi}((\bm{\mu}^\top\bm{x}-y)/\sqrt{\bm{x}^\top\bm{\Sigma}\bm{x}}) \ge 1-\underline{\varepsilon}\\
\Longleftrightarrow &  \bm{\mu}^\top\bm{x}-y \ge {\rm \Phi}^{-1}(1-\underline{\varepsilon})\sqrt{\bm{x}^\top\bm{\Sigma}\bm{x}}\\
\Longleftrightarrow &   \bm{\mu}^\top\bm{x}-y \ge \Vert{\rm \Phi}^{-1}(1-\underline{\varepsilon}) \bm{\Sigma}^{1/2}\bm{x} \Vert_2,
\end{array}
\]
where the first equivalence holds by the linearity of elliptical distributions, the second one is because that $\mathrm{\Phi}(\cdot)$ is non-decreasing, and the last one is due to the fact that $1-\underline{\varepsilon} \ge 0.5$ (which follows from $\underline{\varepsilon} \le \varepsilon < 0.5$). Observe that the optimum is achieved at $y^\star = \bm{\mu}^\top\bm{x} - \Vert{\rm \Phi}^{-1}(1-\underline{\varepsilon}) \bm{\Sigma}^{1/2}\bm{x} \Vert_2$, plugging this in the objective of problem~\eqref{prob:pessimistic chance constrained MDP} then concludes our proof.
\hfill\qed

\subsection{Proofs of Results in Section~\ref{sec: special cases}}
\textit{Proof of Proposition~\ref{prop:pessimistic combined MDP}.} $\;$
By Proposition~$1$ and Proposition~$2$, we have 
\begin{equation*}
\inf_{\mathbb{P}\in\mathcal{F}(\theta)}\mathbb{E}_{\mathbb{P}}[\tilde{\bm{r}}^\top\bm{x}] = -\theta\Vert\bm{x}\Vert_2 + \mathbb{E}_{\hat{\mathbb{P}}}[\tilde{\bm{r}}^\top\bm{x}]
\end{equation*}
and
\begin{equation*}
\inf_{\mathbb{P}\in\mathcal{F'}(\theta)}\mathbb{P}\text{-VaR}_{1-\varepsilon}[\tilde{\bm{r}}^\top\bm{x}] = \bm{\mu}^\top\bm{x} - \Vert{\rm \Phi}^{-1}(1-\underline{\varepsilon}) \bm{\Sigma}^{1/2}\bm{x} \Vert_2
\end{equation*}
with $\underline{\varepsilon}$ as claimed. Substituting the above two equations into \eqref{RR} and rearranging the terms then concludes our proof.
\hfill\qed

\textit{Proof of Proposition~\ref{prop:misocp for srr}.} $\;$
By the definition of $\hat{\mathbb{T}}$, problem~\eqref{prob:srr} can be rewritten as:
\begin{equation*}
\max_{\bm{\pi}\in(\Delta^A)^S}\psi\sum_{i\in[N]}w_i\cdot g(\bm{\pi},\hat{\bm{P}}^i) + (1-\psi)\max_{\eta\in\mathbb{R}}\left\{\eta-\frac{1}{1-\iota}\sum_{i\in[N]}w_i(\eta-g(\bm{\pi},\hat{\bm{P}}^i))^+\right\}.
\end{equation*}
By introducing auxiliary decision variables $\bm{y}\in\mathbb{R}^N$, it can be further reformulated as:
\begin{equation}\label{prob:srr reformulation}
\begin{array}{cll}
\max&\displaystyle\psi\sum_{i\in[N]}w_i\cdot g(\bm{\pi},\hat{\bm{P}}^i) + (1-\psi)\left(\eta-\frac{1}{1-\iota}\sum_{i\in[N]}y_i\right)\\
{\rm s.t.}&y_i\geq w_i(\eta-g(\bm{\pi},\hat{\bm{P}}^i))&\forall i\in[N]\\
&\bm{\pi}\in(\Delta^A)^S, \bm{y}\in\mathbb{R}^N_+,\eta\in\mathbb{R}.
\end{array}
\end{equation}
Here we can express
\begin{equation}\label{prob:srr w times g}
\begin{array}{rcll}
w_i\cdot g(\bm{\pi},\bm{P})=&\max&\bm{\mu}^\top\bm{x}  -\alpha\theta\cdot\Vert\bm{x}\Vert_2 - (1-\alpha)\cdot\Vert{\rm \Phi}^{-1}(1-\underline{\varepsilon}) \bm{\Sigma}^{1/2}\bm{x} \Vert_2\\
&{\rm s.t.}&\displaystyle x_{s,a} = \pi_{s,a}\cdot\sum_{a'\in\mathcal{A}}x_{s,a'}&\forall (s,a)\in\mathcal{S}\times\mathcal{A}\\
&&(\bm{E}-\gamma\cdot\bar{\bm{P}})\bm{x} = w_i\cdot\bm{p}_0\\
&&\bm{x}\in\mathbb{R}^{SA}_+
\end{array}
\end{equation}
as in \cite{lobo2020soft}. We can then, by combining \eqref{prob:srr reformulation} and \eqref{prob:srr w times g}, reformulate problem~\eqref{prob:srr} as:
\begin{equation*}
\begin{array}{cll}
\max&\displaystyle  \psi\sum_{i\in[N]}(\bm{\mu}^\top\bm{x}^i-\alpha\theta\cdot\Vert\bm{x}^i\Vert_2-(1-\alpha)\cdot\Vert{\rm \Phi}^{-1}(1-\underline{\varepsilon}) \bm{\Sigma}^{1/2}\bm{x}^i \Vert_2) + (1-\psi)(\eta-\frac{1}{1-\iota}\sum_{i\in[N]}y_i)\\
{\rm s.t.}&
y_i - w_i \eta \geq \alpha\theta\cdot\Vert\bm{x}^i\Vert_2 + (1-\alpha)\cdot 
\Vert{\rm \Phi}^{-1}(1-\underline{\varepsilon}) \bm{\Sigma}^{1/2}\bm{x}^i \Vert_2-\bm{\mu}^\top\bm{x}^i
&\forall i \in[N]\\
&\displaystyle x^i_{s,a} = \pi_{s,a}\cdot\sum_{a'\in\mathcal{A}}x^i_{s,a'}&\forall i\in[N],(s,a)\in\mathcal{S}\times\mathcal{A}\\
&(\bm{E}-\gamma\cdot\bar{\bm{P}}^{i})\bm{x}^i=w_i\cdot\bm{p}_0&\forall i \in[N] \\
&\bm{\pi}\in(\Delta^A)^S,\eta\in\mathbb{R},\bm{x}^i\in\mathbb{R}^{SA}_+,\bm{y}\in\mathbb{R}^N_+&\forall i\in[N].
\end{array}
\end{equation*}
Now it is sufficient to focus on the second set of constraints
\begin{equation}\label{prob:bilinear constraint}
x^i_{s,a} = \pi_{s,a}\cdot\sum_{a'\in\mathcal{A}}x^i_{s,a'}\;\forall i\in[N], (s,a)\in\mathcal{S}\times\mathcal{A}.
\end{equation}
Since we only consider deterministic policy $\bm{\pi}\in\{0,1\}^{SA}$ and $\sum_{a\in\mathcal{A}}x^i_{s,a}\in[0,w_i/(1-\gamma)]$ (see, \textit{e.g.}, lemma C.10 in \cite{petrik2010optimization}), we have the McCormick relaxation (see, \textit{e.g.}, \cite{petrik2016interpretable}) of \eqref{prob:bilinear constraint} as:
\begin{equation*}
\left\{
\begin{array}{ll}
\displaystyle x^i_{s,a}\leq \sum_{a'\in\mathcal{A}}x^i_{s,a'}\\
\displaystyle x^i_{s,a}\leq \frac{w_i}{1-\gamma}\pi_{s,a}\\
\displaystyle x^i_{s,a}\geq 0\\
\displaystyle x^i_{s,a}\geq \frac{w_i}{1-\gamma}(\pi_{s,a}-1)+\sum_{a'\in\mathcal{A}}x^i_{s,a'}
\end{array}\right.
\end{equation*}
for all $i\in[N], (s,a)\in\mathcal{S}\times\mathcal{A}$. Our conclusion then follows from the fact that the McCormick relaxation is precise when $\pi\in\{0,1\}$ (\textit{i.e.}, the extreme values of the interval $[0,1]$).
\hfill\qed

\subsection{Proofs of Results in Section~\ref{sec:adlpmm}}\label{apd:proof of adlpmm}

\textit{Proof of Proposition~\ref{prop:projection}.} $\;$
By~\eqref{prob:y prox}, it is sufficient to focus on solving ${\rm Proj}_{\bm{B}_{{\ell_{\bm{\Sigma}}(\cdot)}}}(\bm{x})$. By eigenvalue decomposition, we have $\bm{\Sigma}=\bm{G}^\top\bm{D}\bm{G}$\footnotemark\footnotetext{The eigenvalue decomposition here is not counted in the time complexity of the bisection method (or the AD-LPMM algorithm), since this process is carried out for computing $\bm{\Sigma}^{1/2}$ in~\eqref{prob:RR reformulation} (before we solve~\eqref{prob:RR reformulation}).} with $\bm{D}={\rm diag}(d_1, \cdots,d_{SA})$, thus we have:  
\begin{equation*}
\begin{array}{r@{\;\;}c@{\;\;}l}
{\rm Proj}_{\bm{B}_{{\ell_{\bm{\Sigma}}(\cdot)}}}(\bm{x})=&\argmin&\frac{1}{2}\cdot\Vert \bm{v}-\bm{x}\Vert_2^2  \\
&{\rm s.t.}& \displaystyle \bm{v}^\top\bm{G}^\top\bm{D}\bm{G}\bm{v} \leq 1\\
&&\bm{v}\in\mathbb{R}^{SA}.
\end{array}
\end{equation*}
By change of variable $\bm{u}=\bm{G}\bm{v}$ and let $\bm{b}=\bm{G}\bm{x}$, it is sufficient to focus on the equivalent problem:
\begin{equation}\label{prob:bisect u}
\begin{array}{c@{\;\;}l}
\argmin&\frac{1}{2}\cdot\Vert \bm{u}-\bm{b}\Vert_2^2  \\
{\rm s.t.}& \displaystyle \bm{u}^\top\bm{D}\bm{u} \leq 1\\
&\bm{u}\in\mathbb{R}^{SA},
\end{array}
\end{equation}
where we can retrieve $\bm{v}^\star=\bm{G}^\top\bm{u}^\star$. The Lagrangian function of\eqref{prob:bisect u} (with the introduced dual variable $\zeta\in\mathbb{R}_+$) is
\begin{equation*}
L(\bm{u};\zeta) = \frac{1}{2}\cdot\Vert \bm{u}-\bm{b}\Vert_2^2 + \zeta(\bm{u}^\top\bm{D}\bm{u}-1).
\end{equation*}
Since \eqref{prob:bisect u} is a convex optimization problem, the KKT condition is the sufficient condition for the optimality of the primal and dual solutions:
\begin{equation*}
\left\{
\begin{array}{l@{\;\;}l}
\bm{u}^\top\bm{D}\bm{u} \leq 1\\
\zeta \geq 0\\
\zeta(\bm{u}^\top\bm{D}\bm{u}-1) = 0\\
\nabla_{\bm{u}}L(\bm{u};\zeta) = \bm{u}-\bm{b}+2\zeta\cdot\bm{D}\bm{u} = 0,
\end{array}\right.
\end{equation*}
where for $\zeta=0$, we have
\begin{equation*}
\left\{
\begin{array}{l@{\;\;}l}
\bm{u}^\top\bm{D}\bm{u} \leq 1\\
\bm{u}-\bm{b} = 0;
\end{array}\right.
\end{equation*}
while when $\zeta>0$, we have
\begin{equation*}
\left\{
\begin{array}{l@{\;\;}l}
\bm{u}^\top\bm{D}\bm{u} = 1\\
(\bm{I}+2\zeta\cdot\bm{D})\bm{u}-\bm{b} = 0.
\end{array}\right.
\end{equation*}
Therefore, if $\bm{b}^\top\bm{D}\bm{b}\leq 1$, we have $\bm{u}^\star=\bm{b}$; if $\bm{b}^\top\bm{D}\bm{b}> 1$, it is sufficient to solve the equation $g(\zeta)=1$ where
$$
g(\zeta)=\sum_{i\in[SA]}\frac{d_ib_i^2}{(1+2\zeta d_i)^2}.
$$
The function $g$ is monotonically decreasing function on $[0,+\infty)$ and $\lim_{\zeta\rightarrow +\infty}g(\zeta)=0$, thus we can apply the bisection method to search on the interval $[0,\bar{\zeta}]$ (where $\bar{\zeta}:g(\bar{\zeta})\leq 1$ is the upper bound for the search which we provide in Lemma~\ref{lemma:bisect ub}) to locate $\zeta^\star$ and retrieve $u^\star_i=b_i/(1+2\zeta^\star d_i)\; \forall i\in[SA]$. The pseudocode is provided in Algorithm~\ref{alg:bisection}.

The time complexity of solving $\mathfrak{P}_{\bm{y}}(\bm{x},\bm{\xi};c)$ is dominated by the bisection method, which has time complexity $\mathcal{O}(\log(1/\delta'))$. Our conclusion follows from the fact that the computation in each iteraion of the bisection takes time $\mathcal{O}(SA)$.
\hfill\qed

\begin{algorithm}[t]
\caption{Bisection for Problem~\eqref{prob:bisect u} }\label{alg:bisection}
\Input{\textnormal{Desired precision $\delta'$, initial lower bound $\underline{\zeta}\leftarrow 0$ and upper bound $\overline{\zeta}>0$}}\\
\If{$g(0)\leq1$}{$\bm{u} \leftarrow \bm{b}$;}
\Else{
\While{$\vert\overline{\zeta}-\underline{\zeta}\vert\geq\delta'$}{
$\zeta\leftarrow 0.5(\overline{\zeta}+\underline{\zeta})$;

\If { $g(\zeta)>= 1$}{
$\underline{\zeta}\leftarrow \zeta$;
}
\Else {
$\overline{\zeta}\leftarrow \zeta$;
}
}
\For {$i=1,\cdots,SA$}{$u_i=b_i/(1+2\zeta d_i)$;}
}

\Output{\textnormal{Solution $\bm{u}$}}
\end{algorithm}
\begin{lemma}\label{lemma:bisect ub}
\textnormal{The inequality $g(\zeta)\leq1$ holds for all $\zeta\geq(1/(2d_{i''}))(b_{i'}\sqrt{SAd_{i'}}-1)$, where $i'\in\argmax_{i\in[SA]}d_ib_i^2$ and $i''\in\argmin_{i\in[SA]}d_i$}
\end{lemma}
\noindent \textit{Proof.}
Observe that,
\begin{equation*}
\begin{array}{r@{\;\;}l}
g(\zeta)\leq&\displaystyle \sum_{i\in[SA]}\frac{d_{i'}b_{i'}^2}{(1+2\zeta d_i)^2}\\
\leq& \frac{SAd_{i'}b_{i'}^2}{(1+2\zeta d_{i''})^2},\\
\end{array}
\end{equation*}
from which we have 
\begin{equation*}
\frac{SAd_{i'}b_{i'}^2}{(1+2\zeta d_{i''})^2} \leq 1 \Rightarrow g(\zeta) \leq 1.
\end{equation*}
Our conclusion thus follows by rearranging the terms of the inequality on the left-hand side.

\hfill\qed

By Lemma~\ref{lemma:bisect ub}, one can choose $\overline{\zeta} = (1/(2d_{i''}))(b_{i'}\sqrt{SAd_{i'}}-1)$, where $i'\in\argmax_{i\in[SA]}d_ib_i^2$ and $i''\in\argmin_{i\in[SA]}d_i$ for Algorithm~\ref{alg:bisection}.

\textit{Proof of Proposition~\ref{prop:z}.} $\;$
Notice that, it is sufficient to solve the $i^{\rm th}$ subproblem:
\begin{equation*}
\argmin_{z\geq0} \frac{c}{2}z^2 - (cx_i+\mu_i+\eta_i)z = \max\left\{0,\frac{1}{c}(cx_i+\mu_i+\eta_i)\right\}
\end{equation*}
for all $i\in[SA]$, where our conclusion follows.
\hfill\qed

\textit{Proof of Proposition~\ref{prop:x}.} $\;$
By the definition of $\bm{Q}(\cdot,\cdot)$, we have
\begin{equation*}
\begin{array}{r@{\;\;}l}
&\mathfrak{P}_{\bm{x}}(\bm{y},\bm{z},\bm{\lambda},\bm{\xi},\bm{\eta};c,\nu,\hat{\bm{x}})  \\
=&\displaystyle\argmin_{\bm{x}}\;\alpha\theta\cdot\Vert\bm{x}\Vert_2+\bm{x}^\top((\bm{E}-\gamma\cdot\bar{\bm{P}})^\top\bm{\lambda}+\bm{\xi}+\bm{\eta})+\frac{c}{2}\cdot
\left\Vert
\begin{matrix}
(\bm{E}-\gamma\cdot\bar{\bm{P}})(\bm{x}-\hat{\bm{x}})+(\bm{E}-\gamma\cdot\bar{\bm{P}})\hat{\bm{x}}-\bm{p}_0\\
\bm{x}-\hat{\bm{x}}+\hat{\bm{x}}-\bm{y}\\
\bm{x}-\hat{\bm{x}}+\hat{\bm{x}}-\bm{z}
\end{matrix}\right\Vert_2^2\\
&+\frac{1}{2}\cdot
\ell^2_{\bm{Q}(c,\nu)}(\bm{x}-\hat{\bm{x}})\\
=&\displaystyle\argmin_{\bm{x}}\;\alpha\theta\cdot\Vert\bm{x}\Vert_2+\bm{x}^\top((\bm{E}-\gamma\cdot\bar{\bm{P}})^\top\bm{\lambda}+\bm{\xi}+\bm{\eta})+\frac{c}{2}\cdot
\left\Vert
\begin{matrix}
(\bm{E}-\gamma\cdot\bar{\bm{P}})(\bm{x}-\hat{\bm{x}})\\
\bm{x}-\hat{\bm{x}}\\
\bm{x}-\hat{\bm{x}}
\end{matrix}\right\Vert_2^2\\
&+c\cdot\bm{x}^\top\left((\bm{E}-\gamma\cdot\bar{\bm{P}})^\top\left((\bm{E}-\gamma\cdot\bar{\bm{P}})\hat{\bm{x}}-\bm{p}_0\right)+2\cdot\hat{\bm{x}}-\bm{y}-\bm{z}\right)
+\frac{1}{2}\cdot
\ell^2_{\bm{Q}(c,\nu)}(\bm{x}-\hat{\bm{x}})\\
=&\displaystyle\argmin_{\bm{x}}\;\frac{\alpha\theta}{c\nu}\cdot\Vert\bm{x}\Vert_2+\bm{x}^\top\bm{w}+\frac{1}{2}\cdot\Vert\bm{x}-\hat{\bm{x}}\Vert^2_2\\
=&\displaystyle\argmin_{\bm{x}}\;\frac{\alpha\theta}{c\nu}\cdot\Vert\bm{x}\Vert_2+\frac{1}{2}\cdot\Vert\bm{x}-(\hat{\bm{x}}-\bm{w})\Vert^2_2\\
=&\left(1-\frac{\frac{\alpha\theta}{c\nu}}{\max\{\Vert\bm{w}\Vert_2,\frac{\alpha\theta}{c\nu}\}}\right)\cdot(\hat{\bm{x}}-\bm{w})
\end{array}
\end{equation*}
where we denote $\bm{w}=\frac{1}{c\nu}\cdot\left(\left(\bm{E}-\gamma\cdot\bar{\bm{P}}\right)^\top\bm{\lambda}+\bm{\xi}+\bm{\eta}\right)+\frac{1}{\nu}\cdot\left( \left(\bm{E}-\gamma\cdot\bar{\bm{P}}\right)^\top\left(\left(\bm{E}-\gamma\cdot\bar{\bm{P}}\right)\hat{\bm{x}}-\bm{p}_0\right)+2\cdot\hat{\bm{x}}-\bm{y}-\bm{z} \right)$, and the last equality holds by, \textit{e.g.}, example~$6.1.9$ in \cite{beck2017first}.

The computation time is dominated by computing $\Vert\bm{w}\Vert_2$, which is $\mathcal{O}(SA)$.
\hfill\qed

\section{Evaluation of VaR and CVaR of Student's \textit{t}-Distribution}\label{apd:t distribution}
The VaR of a Student's $t$-distribution with threshold $\varepsilon$ is in fact the lower-$\varepsilon$ percentile of its probability density function (PDF), which can be looked up in table in, \textit{e.g.}, \cite{hogg1995introduction} (under some common values of $\varepsilon<0.5$). We provide the calculation of CVaR as follows (with degree of freedom $\delta>1$ and $v:=\mathbb{P}_{t\text{-}{\rm dist}}\text{-}{\rm VaR}_{\varepsilon}(\tilde{r})$ assumed known): 
\begin{equation*}
\begin{array}{rcl}
\mathbb{P}_{t\text{-}{\rm dist}}\text{-}{\rm CVaR}_{\varepsilon}(\tilde{r})& =&\frac{1}{\varepsilon}\cdot\frac{\Gamma(\frac{\delta+1}{2})}{(\pi \delta)^{\frac{1}{2}}\Gamma(\frac{\delta}{2})}\int^v_{-\infty}\frac{r}{(1+\frac{r^2}{\delta})^{\frac{\delta+1}{2}}}{\rm d}r\\
&=&\frac{1}{\varepsilon}\cdot\frac{\delta^{\frac{1}{2}}\cdot\Gamma(\frac{\delta+1}{2})}{2\pi ^{\frac{1}{2}}\Gamma(\frac{\delta}{2})}\int^{1+\frac{v^2}{\delta}}_{-\infty}u^{-\frac{k+1}{2}}{\rm d}u\\
&=&-\frac{\delta^{\frac{1}{2}}\cdot\Gamma(\frac{\delta+1}{2})}{\varepsilon\pi ^{\frac{1}{2}}(\delta-1)\Gamma(\frac{\delta}{2})}\cdot \left(1+\frac{v^2}{\delta}\right)^{-\frac{k-1}{2}},
\end{array}
\end{equation*}
where the first equality follows from the definition of the CVaR and the PDF of the $t$-distribution herein, the second equality holds by the technique of integration by substitution.

\section{Preliminaries on Elliptical Distributions}\label{apd:elliptical distributions}

The probability density distribution of an elliptical reference distribution $\mathbb{P}_{(\bm{\mu},\bm{\Sigma},g)}$ is given by
$$
f(\bm{r})=k \cdot g\left(\frac{1}{2}(\bm{r}-\bm{\mu})^\top \bm{\Sigma}^{-1}(\bm{r}-\bm{\mu})\right),
$$
where $k$ is a positive normalization scalar, $\bm{\mu}$ is a mean vector, $\bm{\Sigma}$ is a positive definite matrix and $g$ is a generating function. Elliptical distribution is a broad family of distributions that includes for example, the multivariate normal distribution, multivariate $t$-distribution and multivariate logistic distribution, as special cases. One notable property of the elliptical distribution is the linearity: any linear combination of elliptically distributed random variables still follows an elliptical distribution. That is, for any random vector $\tilde{\bm{r}} \sim \mathbb{P}_{(\bm{\mu}, \bm{\Sigma}, g)}$, it holds that $\tilde{\bm{r}}^\top \bm{x} \sim $ $\mathbb{P}_{\left(\mu_{\bm{x}}, \sigma_{\bm{x}}^2, g\right)}$ with $\mu_{\bm{x}}=\bm{\mu}^\top \bm{x}$ and $\sigma_{\bm{x}}=\sqrt{\bm{x}^{\top} \bm{\Sigma} \bm{x}}$. Indeed, we can express the combination as $\tilde{\bm{r}}^\top \bm{x}=\mu_{\bm{x}}+\sigma_{\bm{x}} \tilde{z}$, where $\tilde{z} \sim \mathbb{P}_{(0,1, g)}$ is a standard elliptically distributed random variable whose probability density function and cumulative distribution function are
$
\phi(z)=k \cdot g\left(z^{2} / 2\right)
$
and 
$
\mathrm{\Phi}(x)=\int_{-\infty}^{x} k \cdot g (z^2/2) \mathrm{d} z,
$
respectively. For a concrete example we take a closer look at a standard normal distribution, for which the normalization scalar and generating function are $k=1 / \sqrt{2 \pi}$ and $g(x)=\exp(-x)$, respectively.

\section{Distributionally Optimistic MDPs}\label{apd:optimistic MDPs}

In contrast to the robust model, sometimes the decision maker prefers exploration over exploitation if she would like to learn more information about the MDP. As such, we could instead adopt an optimistic counterpart where we focus on the best case, motivating the following distributionally optimistic MDP:
\begin{equation}\label{prob:optimistic MDP}
\ell_{\rm O}(\theta) = 
\max_{\bm{x} \in \mathcal{X}} \; \sup_{\mathbb{P}\in\mathcal{F}(\theta)}\mathbb{E}_{\mathbb{P}}[\tilde{\bm{r}}^\top\bm{x}].
\end{equation}
In contrast to the robust case, here our decision depends instead on the best possible (expected) outcome, which exactly embodies optimism. We summarize the reformulation of~\eqref{prob:optimistic MDP} as follows.


\begin{proposition}\label{prop:DOMDP}
The distributionally optimistic MDP~\eqref{prob:optimistic MDP} is equivalent to an optimization problem
\begin{equation}\label{prob:optimistic MDP reformulation}
\nonumber
\ell_{\rm O}(\theta) = 
\max_{\bm{x} \in \mathcal{X}} \; \mathbb{E}_{\hat{\mathbb{P}}}[\tilde{\bm{r}}^\top\bm{x}]+ \theta \Vert\bm{x}\Vert_*.
\end{equation}
\end{proposition}
\noindent \textit{Proof.}
It is sufficient to rewrite the objective of~\eqref{prob:optimistic MDP} as follows:
\[
\sup_{\mathbb{P}\in\mathcal{F}(\theta)} \mathbb{E}_\mathbb{P}[\tilde{\bm{r}}^\top\bm{x}]
= - \inf_{\mathbb{P}\in\mathcal{F}(\theta)} \mathbb{E}_\mathbb{P}[-\tilde{\bm{r}}^\top\bm{x}] 
= -(\mathbb{E}_{\hat{\mathbb{P}}}[-\tilde{\bm{r}}^\top\bm{x}]-\theta\Vert\bm{x}\Vert_*) 
= \mathbb{E}_{\hat{\mathbb{P}}}[\tilde{\bm{r}}^\top\bm{x}] + \theta\Vert\bm{x}\Vert_*,
\]
where the second identity follows similar lines as in the proof of Proposition~$1$.
\hfill\qed

The reformulation in Proposition~\ref{prop:DOMDP} is a reverse conic program that is, in general, non-convex. However, it can be recast as a mixed-integer linear program, provided that $\|\cdot\|_*$ is the commonly used $L_1$-norm or $L_{\infty}$-norm. Such a mixed-integer linear program can be solved by the state-of-the-art approaches.

\section{Distributionally Optimistic Chance-Constrained Model}\label{apd:optimistic CC}
In a distributionally optimistic chance-constrained MDP model, where we focus on the best case that with high probability, the reward is no smaller than some lower bound that we maximize. Formally, the distributionally optimistic chance-constrained MDP model is formulated as follows:
\begin{equation}\label{prob:optimistic chance constrained MDP}
\ell_{\rm DOCC}(\theta, \varepsilon) = \left\{
\begin{array}{c@{\;\;}l@{\;\;}l}
\max &~ \displaystyle y\\
{\rm s.t.} & \displaystyle  \sup_{\mathbb{P} \in \mathcal{F}(\theta)}\mathbb{P}[\tilde{\bm{r}}^\top\bm{x}\ge y] \ge 1 - \varepsilon \\
&\displaystyle  ~\bm{x} \in \mathcal{X},\; y \in \mathbb{R}.
\end{array}
\right.
\end{equation}
The optimistic chance-constrained model~\eqref{prob:optimistic chance constrained MDP} is also equivalent to a nominal chance-constrained model, however, at a less risky level. Before formally establishing this argument, two lemmas are introduced as follows.

\begin{lemma}\label{lemma:largest probability}
The worst (largest) probability of the random vector $\tilde{\bm{r}}$ attaining a value in the set $\mathcal{R}$,
\begin{equation}\label{prob:largest probability}
\sup _{\mathbb{P} \in \mathcal{F}(\theta)}\mathbb{P}[\tilde{\bm{r}} \in \mathcal{R}],
\end{equation}
is equivalent to 
\begin{equation*}
\min _{\lambda \ge 0}\left\{\lambda \theta+\int_{\bm{r} \in \mathbb{R}^{SA}}(\lambda \cdot \mathbf{dist}(\bm{r}, \mathcal{R})-1)^{-} \mathrm{d} \hat{\mathbb{P}}\bm{r}\right\}.
\end{equation*}
Here, we use $\mathbf{dist}(\bm{r},\mathcal{R}) = \inf\{\Vert \bm{r}-\hat{\bm{r}}\Vert ~\vert~ \hat{\bm{r}} \in \mathcal{R}   \}$ to denote the distance from the vector $\bm{r} \in \mathbb{R}^{SA}$ to the set $\mathcal{R} \subseteq \mathbb{R}^{SA}$.
\end{lemma}

\noindent \textit{Proof.}
Using theorem 1 in \cite{gao2016distributionally} or theorem 1 in \cite{blanchet2019quantifying}, the uncertainty quantification problem~\eqref{prob:largest probability} is equal to 
\begin{equation}\label{prob:largest probability dual 1}
\min _{\lambda \ge 0}\left\{\lambda \theta-\int_{\bm{r} \in \mathbb{R}^{SA}} \inf _{\bm{w} \in \mathbb{R}^{SA}}\{\lambda\Vert\bm{w}-\bm{r}\Vert-\mathbb{I}[\bm{w} \in \mathcal{R}]\} \mathrm{d}\hat{\mathbb{P}} \bm{r}\right\},
\end{equation}
where $\mathbb{I}$ is the 0-1 indicator function. Consider the second term in the objective of the above minimization problem, we have
\begin{equation}\label{prob:largest probability dual second term}
\inf _{\bm{w} \in \mathbb{R}^{SA}}\{\lambda\Vert\bm{w}-\bm{r}\Vert-\mathbb{I}[\bm{w} \in \mathcal{R}]\}=-(\lambda \cdot \mathbf{dist}(\bm{r}, \mathcal{R})-1)^{-}.
\end{equation}
Indeed, if $\bm{r}\in\mathcal{R}$ (for which, $\mathbf{dist}(\bm{r},\mathcal{R}) = 0$), then by choosing $\bm{w} = \bm{v}$, it holds that
\begin{equation*}
\inf _{\bm{w} \in \mathbb{R}^{SA}}\{\lambda\Vert\bm{w}-\bm{r}\Vert-\mathbb{I}[\bm{w} \in \mathcal{R}]\}=-1=-(\lambda \cdot \mathbf{dist}(\bm{r}, \mathcal{R})-1);
\end{equation*}
whereas if $\bm{r} \notin \mathcal{R}$, then it holds that 
\begin{equation*}
\begin{array}{r@{\;\;}l}
\displaystyle \inf _{\bm{w} \in \mathbb{R}^{SA}}\{\lambda\Vert\bm{w}-\bm{r}\Vert-\mathbb{I}[\bm{w} \in \mathcal{R}]\}= & \displaystyle \min \left\{\inf _{\bm{w} \in \mathcal{R}}\{\lambda\Vert\bm{w}-\bm{r}\Vert-1\}, \inf _{\bm{w} \notin \mathcal{R}} \lambda\Vert\bm{w}-\bm{r}\Vert\right\} \\[3mm]
= & \displaystyle\min \left\{\inf _{\bm{w} \in \mathcal{R}                }\{\lambda\Vert\bm{w}-\bm{r}\Vert-1\}, 0\right\}\\[3mm]
= & -(\lambda \cdot \mathbf{dist}(\bm{r}, \mathcal{R})-1)^{-}.
\end{array}
\end{equation*}
Plugging expression (\ref{prob:largest probability dual second term}) into problem~\eqref{prob:largest probability dual 1} gives the desired result, which, by proposition 3 in \cite{gao2016distributionally}, holds regardless of whether $\mathcal{R}$ is open or closed.
\hfill\qed

\begin{lemma}\label{lemma:optimistic chance constraint}
The distributionally optimistic chance constraint 
\begin{equation}\label{prob:optimistic chance constraint 2}
\inf _{\mathbb{P} \in \mathcal{F}(\theta)} \mathbb{P}[\tilde{\bm{r}} \in \mathcal{R}] \le \varepsilon
\end{equation}
with a risk threshold $\varepsilon \in (0,1)$ is satisfiable if and only if 
\begin{equation}\label{prob:optimistic chance constraint cvar}
\nonumber
\mathbb{P}\text{-}\mathrm{CVaR}_{\varepsilon}[-\mathbf{dist}(\tilde{\bm{r}}, \bar{\mathcal{R}})] \ge-\frac{\theta}{1-\varepsilon},
\end{equation}
where $\bar{\mathcal{R}}=\mathbb{R}^{SA} \; \backslash \; \mathcal{R}$ is the complement of the set of undesired events $\mathcal{R}$.

\end{lemma}

\noindent \textit{Proof.}
We first re-express (\ref{prob:optimistic chance constraint 2}) as
\begin{equation*}
\sup _{\mathbb{P} \in \mathcal{F}(\theta)} \mathbb{P}[\tilde{\bm{r}} \in \bar{\mathcal{R}}] \ge 1-\varepsilon.
\end{equation*}
Using Lemma~\ref{lemma:largest probability}, the above constraint is equivalent to
\begin{equation}\label{prob:optimistic chance constraint 3}
\min _{\lambda \ge 0}\left\{\lambda \theta+\int_{\bm{r} \in \mathbb{R}^{SA}}(\lambda \cdot \mathbf{dist}(\bm{r}, \bar{\mathcal{R}})-1)^{-} \mathrm{d}\hat{\mathbb{P}} \bm{r}\right\} \ge 1-\varepsilon.
\end{equation}
The left-hand side problem can be presented by 
\begin{equation*}
\min \left\{\min _{\lambda>0}\left\{\lambda \theta+\int_{\bm{r} \in \mathbb{R}^{SA}}(\lambda \cdot \mathbf{dist}(\bm{r}, \bar{\mathcal{R}})-1)^{-} \mathrm{d} \hat{\mathbb{P}}\bm{r}\right\}, 1\right\}.
\end{equation*}
Since $1 \ge 1 - \varepsilon$, the above re-expression implies that constraint (\ref{prob:optimistic chance constraint 3}) is equivalent to 
\begin{equation*}
\min _{\lambda>0}\left\{\lambda \theta+\int_{\bm{r} \in \mathbb{R}^{SA}}(\lambda \cdot \mathbf{dist}(\bm{r}, \bar{\mathcal{R}})-1)^{-} \mathrm{d} \hat{\mathbb{P}}\bm{r}\right\} \ge 1-\varepsilon.
\end{equation*}
Multiplying both sides by $(\lambda(1-\varepsilon))^{-1}>0$, we arrive at
\begin{equation*}
\min _{\tau<0}\left\{\frac{1}{1-\varepsilon} \int_{\bm{r} \in \mathbb{R}^{SA}}(-\mathbf{dist}(\bm{r}, \bar{\mathcal{R}})-\tau)^{+} \mathrm{d} \hat{\mathbb{P}}\bm{r}+\tau\right\} \ge-\frac{\theta}{1-\varepsilon},
\end{equation*}
which, together with the fact
\begin{equation*}
\min _{\tau \ge 0}\left\{\frac{1}{1-\varepsilon} \int_{\bm{r} \in \mathbb{R}^{SA}}(-\mathbf{dist}(\bm{r}, \bar{\mathcal{R}})-\tau)^{+} \mathrm{d} \hat{\mathbb{P}}\bm{r}+\tau\right\} \ge 0 \ge-\frac{\theta}{1-\varepsilon},
\end{equation*}
is equivalent to
\begin{equation*}
\min _{\tau \in \mathbb{R}}\left\{\frac{1}{1-\varepsilon} \int_{\bm{r} \in \mathbb{R}^{SA}}(-\mathbf{dist}(\bm{r}, \bar{\mathcal{R}})-\tau)^{+} \mathrm{d} \hat{\mathbb{P}}\bm{r}+\tau\right\} \ge-\frac{\theta}{1-\varepsilon},
\end{equation*}
where the left-hand side is essentially $\hat{\mathbb{P}}$-$\mathrm{CVaR}_{\varepsilon}[-\mathbf{dist}(\tilde{\bm{r}}, \bar{\mathcal{R}})]$.
\hfill\qed

Now we are ready to establish the equivalence between the chance-constrained model and its optimistic counterpart (with an adjusted risk threshold).

\begin{lemma}\label{lemma:optimistic robust chance constraint}
Suppose in the Wasserstein ambiguity set \eqref{set:Wasserstein ball}, the reference distribution is an elliptical distribution $\hat{\mathbb{P}} = \mathbb{P}_{(\bm{\mu}, \bm{\Sigma}, g)}$ and the Wasserstein distance is equipped with a Mahalanobis norm associated with the positive definite matrix $\bm{\Sigma}$. The distributionally optimistic robust chance constraint
\begin{equation}\label{prob:optimistic chance constraint}
\nonumber
\exists \; \mathbb{P} \in \mathcal{F}(\theta): \mathbb{P}[\tilde{\bm{r}}^\top\bm{x} \ge y] \ge 1-\varepsilon
\end{equation}
is satisfiable if and only if
$
\mathbb{P}_{(\bm{\mu}, \bm{\Sigma}, g)}[\tilde{\bm{r}}^\top\bm{x} \ge y] \ge 1-\bar{\varepsilon},
$
where $\bar{\varepsilon}=1-\mathrm{\Phi}(\underline{\eta}^\star)\ge\varepsilon$ with $\underline{\eta}^\star$ being the smallest $\eta \le \mathrm{\Phi}^{-1}(1-\varepsilon)$ that satisfies
$
\eta({\rm \Phi}(\eta)-(1-\varepsilon))+\int_{\eta^{2} / 2}^{\left({\rm \Phi}^{-1}(1-\varepsilon)\right)^{2} / 2} k g(z) \mathrm{d} z \le \theta.
$
\end{lemma}

\noindent \textit{Proof.}
We first look at the individual distributionally optimistic robust chance constraint
\begin{equation*}
\exists\; \mathbb{P} \in \mathcal{F}(\theta): \mathbb{P}[\tilde{\bm{r}}^\top\bm{x}\ge y] \ge 1-\varepsilon
\end{equation*}
for some generic coefficient vector $\bm{x} \in \mathbb{R}^{SA}$. The above chance constraint is equivalent to
\begin{equation*}
\sup _{\mathbb{P} \in \mathcal{F}(\theta)} \mathbb{P}[\tilde{\bm{r}}^\top\bm{x}\ge y] \ge 1-\varepsilon  ~\Longleftrightarrow~ \sup _{\mathbb{P} \in \mathcal{F}(\theta)} \mathbb{P}[\tilde{\bm{r}}^\top\bm{x}> y] \ge 1-\varepsilon ~\Longleftrightarrow~ \inf _{\mathbb{P} \in \mathcal{F}(\theta)} \mathbb{P}[\tilde{\bm{r}}^\top\bm{x} \le y] \le \varepsilon,
\end{equation*}
where for the first equivalence, by using proposition 3 in \cite{gao2016distributionally}
, it is indifferent
to replace the strict inequality with a weak one. Exploring the definition of VaR, we note that
\begin{equation*}
\inf _{\mathbb{P} \in \mathcal{F}(\theta)} \mathbb{P}[\tilde{\bm{r}}^\top\bm{x} \le y] \le \varepsilon \Longleftrightarrow \inf _{\mathbb{P} \in \mathcal{F}(\theta)} \mathbb{P}\text{-}\mathrm{VaR}_{1-\varepsilon}[y-\tilde{\bm{r}}^\top\bm{x}] \le 0.
\end{equation*}
Hence, with the translation invariance of VaR, it is sufficient to show that
\begin{equation}\label{prob:optimistic chance constraint 4}
\inf_{\mathbb{P} \in \mathcal{F}(\theta)} \mathbb{P}\text{-}\mathrm{VaR}_{1-\varepsilon}[-\tilde{\bm{r}}^\top\bm{x}] \triangleq \inf _{v \in \mathbb{R}}\left\{v ~|~ \inf _{\mathbb{P} \in \mathcal{F}(\theta)} \mathbb{P}[-\tilde{\bm{r}}^\top\bm{x}>v] \le \varepsilon  \right\}.
\end{equation}
By Lemma~\ref{lemma:optimistic chance constraint} and the assumption of Mahalanobis norm, we have
\begin{equation*}
\begin{array}{c@{\;\;}l}
\displaystyle \inf _{\mathbb{P} \in \mathcal{F}(\theta)} \mathbb{P}\left[-\tilde{\bm{r}}^\top\bm{x}>v\right] \le \varepsilon &  \displaystyle \Longleftrightarrow \mathbb{P}_{(\bm{\mu}, \bm{\Sigma}, g)}\text{-}\mathrm{CVaR}_{\varepsilon}[-\mathbf{dist}(\tilde{\bm{r}}, \bar{\mathcal{R}})] \geq-\frac{\theta}{1-\varepsilon} \\[3mm]
&\displaystyle \Longleftrightarrow -\mathbb{P}_{(\bm{\mu}, \bm{\Sigma}, g)}\text{-}\mathrm{CVaR}_{\varepsilon}[-(-\tilde{\bm{r}}^\top\bm{x}-v)^{+}] \le \frac{\theta\|\bm{x}\|_{\bm{\Sigma}^{-1}}}{1-\varepsilon},
\end{array}
\end{equation*}
where $\bar{\mathcal{R}}=\left\{\bm{r} ~|~ -\bm{r}^\top\bm{x} \le v\right\}$ and we leverage the closed form solution 
$$
\mathbf{dist}(\tilde{\bm{r}}, \bar{\mathcal{R}})=\left(-\tilde{\bm{r}}^\top\bm{x}-v\right)^{+} /\|\bm{x}\|_{\bm{\Sigma}^{-1}};
$$
see, \textit{e.g.}, lemma 2 in \cite{chen2018data}.

Let $\mathbb{P}_\mathrm{S} = \mathbb{P}_{(\bm{\mu},\bm{\Sigma},g)}$ for simplicity. By the property of elliptical distribution, for $\tilde{\bm{r}} \sim \mathbb{P}_\mathrm{S}$ and any real vector $\bm{x}$, we have $-\tilde{\bm{r}}^\top\bm{x} \sim \mathbb{P}_{(\mu_\mathrm{S},\sigma_\mathrm{S}^2,g)} = \mathbb{P}_{(-\bm{\mu}^\top\bm{x},\bm{x}^\top\bm{\Sigma}\bm{x},g)}$. We denote its probability density function as
\begin{equation*}
h(z) = \frac{k}{\sigma_\mathrm{S}} \cdot g\left(\frac{\left(z-\mu_\mathrm{S}\right)^{2}}{2 \sigma_\mathrm{S}^{2}}\right).
\end{equation*}
The left-hand side of the constraint can be further transformed as
\begin{equation*}
\begin{array}{c@{\;\;}l}
& \displaystyle -\mathbb{P}_\mathrm{S}\text{-}\mathrm{CVaR}_{\varepsilon}[-(-\tilde{\bm{r}}^\top\bm{x}-v)^{+}]\\[3mm]
= &  \displaystyle  -\mathbb{E}_{\mathbb{P}_\mathrm{S}}[-(-\tilde{\bm{r}}^\top\bm{x}-v)^+ ~\vert~ -(-\tilde{\bm{r}}^\top\bm{x}-v)^+ \ge \mathbb{P}_\mathrm{S}\text{-VaR}_{\varepsilon}[-(-\tilde{\bm{r}}^\top\bm{x}-v)^+]] \\[3mm]
= &  \displaystyle -\frac{1}{1-\varepsilon} \int_{-\infty}^{\sup\{ z \vert -(z-v)^+\ge\mathbb{P}_\mathrm{S}\text{-VaR}_{\varepsilon}[-(-\tilde{\bm{r}}^\top\bm{x}-v)^{+}]\}}-(z-v)^+h(z)\mathrm{d}z \\[3mm]
= &  \displaystyle \frac{1}{1-\varepsilon} \int_{v}^{\sup\{ z \vert -(z-v)^+\ge\mathbb{P}_\mathrm{S}\text{-VaR}_{\varepsilon}[-(-\tilde{\bm{r}}^\top\bm{x}-v)^{+}]\}}(z-v)h(z)\mathrm{d}z \\[3mm]
= &  \displaystyle \frac{1}{1-\varepsilon} \int_{v}^{ \mathbb{P}_\mathrm{S}\text{-VaR}_{1-\varepsilon}[-\tilde{\bm{r}}^\top\bm{x}]  }(z-v)h(z)\mathrm{d}z, 
\end{array}
\end{equation*}
in which the last equality holds from 
\begin{equation*}
\begin{array}{c@{\;\;}l}
& \displaystyle \sup\{ z ~\vert~ -(z-v)^+\ge\mathbb{P}_\mathrm{S}\text{-VaR}_{\varepsilon}[-(-\tilde{\bm{r}}^\top\bm{x}-v)^{+}]\}\\[3mm]
= &  \displaystyle  \sup\{ z ~|~ \min\{v-z,0\} \ge \mathbb{P}_\mathrm{S}\text{-VaR}_{\varepsilon}[\min\{ v + \tilde{\bm{r}}^\top\bm{x},0 \}] \} \\[3mm]
= &  \displaystyle  \sup\{ z ~|~ \min\{-z,-v\} \ge \mathbb{P}_\mathrm{S}\text{-VaR}_{\varepsilon}[\min\{  \tilde{\bm{r}}^\top\bm{x},-v \}] \} \\[3mm]
= &  \displaystyle  \sup\{ z ~|~ -z \ge \mathbb{P}_\mathrm{S}\text{-VaR}_{\varepsilon}[\min\{  \tilde{\bm{r}}^\top\bm{x},-v \}] \} \\[3mm]
= &  \displaystyle  \sup\{ z ~|~ z \le \mathbb{P}_\mathrm{S}\text{-VaR}_{1-\varepsilon}[\max\{   -\tilde{\bm{r}}^\top\bm{x}, v \}] \} \\[3mm]
= &  \displaystyle  \sup\{ z ~|~ z \le \mathbb{P}_\mathrm{S}\text{-VaR}_{1-\varepsilon}[   -\tilde{\bm{r}}^\top\bm{x}] \} \\[3mm]
= &  \displaystyle  \mathbb{P}_\mathrm{S}\text{-VaR}_{1-\varepsilon}[  -\tilde{\bm{r}}^\top\bm{x}].
\end{array}
\end{equation*}
Here, the second equality is due to the translation invariance of VaR, the third one follows from $-v \ge \mathbb{P}_\mathrm{S}\text{-VaR}_{\varepsilon}[\min\{  \tilde{\bm{r}}^\top\bm{x},-v\}]$, the fifth one is because that for any $\varepsilon \in (0,1)$, the distributionally optimistic robust VaR satisfies
\begin{equation}\label{prob:optimistic chance constraint second constraint}
v = \inf _{\mathbb{P} \in \mathcal{F}(\theta)} \mathbb{P}\text{-}\mathrm{VaR}_{1-\varepsilon}[-\tilde{\bm{r}}^\top\bm{x}] \le 
\mathbb{P}_\mathrm{S}\text{-}\mathrm{VaR}_{1-\varepsilon}[-\tilde{\bm{r}}^\top\bm{x}],
\end{equation}
thus the $1-\varepsilon$ quantiles of $-\tilde{\bm{r}}^\top\bm{x}$ and $\max\{ -\tilde{\bm{r}}^\top\bm{x},v \}$ coincide.

Let us denote $q_{1-\varepsilon}=\mathbb{P}_\mathrm{S}\text{-}\mathrm{VaR}_{1-\varepsilon}[-\tilde{\bm{r}}^\top\bm{x}]$, which, by its definition, satisfies
\begin{equation*}
\frac{q_{1-\varepsilon}-\mu_\mathrm{S}}{\sigma_\mathrm{S}}=\mathbb{P}_\mathrm{S}\text{-}\mathrm{VaR}_{1-\varepsilon}\left[\frac{-\tilde{\bm{r}}^\top\bm{x}-\mu_\mathrm{S}}{\sigma_\mathrm{S}}\right]=\mathbb{P}_{(0,1,g)}^0\text{-VaR}_{1-\varepsilon}[\tilde{z}]=\mathrm{\Phi}^{-1}(1-\varepsilon),
\end{equation*}
Here, the first equality holds for the translation invariance and the positive homogeneity of VaR, while the last one follows from the definition of VaR under the standard elliptical distribution $\mathbb{P}_{(0,1,g)}$.

Following the last reformulation of the constraint, we further have
\begin{equation*}
\frac{1}{1-\varepsilon} \int_{v}^{ q_{1-\varepsilon}  }(z-v)h(z) \mathrm{d}z
= \frac{1}{1-\varepsilon} \int^{q_{1-\varepsilon}}_{v} z \cdot \frac{k}{\sigma_\mathrm{S}} \cdot g\left(\frac{\left(z-\mu_\mathrm{S}\right)^{2}}{2 \sigma_\mathrm{S}^{2}}\right) \mathrm{d} z - \frac{v}{1-\varepsilon} \int^{q_{1-\varepsilon}}_{v} \frac{k}{\sigma_\mathrm{S}} \cdot g\left(\frac{\left(z-\mu_\mathrm{S}\right)^{2}}{2 \sigma_\mathrm{S}^{2}}\right) \mathrm{d} z.
\end{equation*}
For its first component, we have
\begin{equation*}
\begin{array}{c@{\;\;}l}
& \displaystyle \frac{1}{1-\varepsilon} \int^{q_{1-\varepsilon}}_{v} z \cdot \frac{k}{\sigma_\mathrm{S}} \cdot g\left(\frac{\left(z-\mu_\mathrm{S}\right)^{2}}{2 \sigma_\mathrm{S}^{2}}\right) \mathrm{d} z \\[3mm]
= &  \displaystyle  \frac{1}{1-\varepsilon} \int^{q_{1-\varepsilon}}_{v} \frac{z-\mu_\mathrm{S}}{\sigma_\mathrm{S}} \cdot k \cdot g\left(\frac{\left(z-\mu_\mathrm{S}\right)^{2}}{2 \sigma_\mathrm{S}^{2}}\right) \mathrm{d} z + \frac{1}{1-\varepsilon} \int^{q_{1-\varepsilon}}_{v}  \frac{\mu_\mathrm{S}}{\sigma_\mathrm{S}} \cdot k \cdot g\left(\frac{\left(z-\mu_\mathrm{S}\right)^{2}}{2 \sigma_\mathrm{S}^{2}}\right) \mathrm{d} z \\[3mm]
= &  \displaystyle  \frac{\sigma_\mathrm{S}}{1-\varepsilon}\ \int^{q_{1-\varepsilon}}_{v} \frac{z-\mu_\mathrm{S}}{\sigma_\mathrm{S}} \cdot k \cdot g\left(\frac{\left(z-\mu_\mathrm{S}\right)^{2}}{2 \sigma_\mathrm{S}^{2}}\right) \mathrm{d} \left(\frac{z-\mu_\mathrm{S}}{\sigma_\mathrm{S}}\right) + \\[3mm]
& \displaystyle \frac{\mu_\mathrm{S}}{1-\varepsilon} \left(\mathrm{\Phi}\left(\frac{q_{1-\varepsilon}-\mu_\mathrm{S}}{\sigma_\mathrm{S}}\right) - \mathrm{\Phi}\left(\frac{v-\mu_\mathrm{S}}{\sigma_\mathrm{S}}\right) \right) \\[3mm]
= & \displaystyle  \frac{\sigma_\mathrm{S}}{1-\varepsilon} \int^{\frac{q_{1-\varepsilon}-\mu_\mathrm{S}}{\sigma_\mathrm{S}}}_{\frac{v-\mu_\mathrm{S}}{\sigma_\mathrm{S}}} t \cdot k \cdot g\left(\frac{t^2}{2}\right) \mathrm{d} \left(\frac{z-\mu_\mathrm{S}}{\sigma_\mathrm{S}}\right) + \frac{\mu_\mathrm{S}}{1-\varepsilon} \left(\mathrm{\Phi}\left(\frac{q_{1-\varepsilon}-\mu_\mathrm{S}}{\sigma_\mathrm{S}}\right) - \mathrm{\Phi}\left(\frac{v-\mu_\mathrm{S}}{\sigma_\mathrm{S}}\right) \right) \\[3mm]
= & \displaystyle  \frac{\sigma_\mathrm{S}}{1-\varepsilon} \int^{\frac{(q_{1-\varepsilon}-\mu_\mathrm{S})^2}{2\sigma_\mathrm{S}^2}}_{\frac{(v-\mu_\mathrm{S})^2}{2 \sigma_\mathrm{S}^2}}   k \cdot g(z) \mathrm{d} z + \frac{\mu_\mathrm{S}}{1-\varepsilon} \left(\mathrm{\Phi}\left(\frac{q_{1-\varepsilon}-\mu_\mathrm{S}}{\sigma_\mathrm{S}}\right) - \mathrm{\Phi}\left(\frac{v-\mu_\mathrm{S}}{\sigma_\mathrm{S}}\right) \right),
\end{array}
\end{equation*}
while for the second component, it holds that
\begin{equation*}
\begin{array}{r@{\;\;}c@{\;\;}l}
\displaystyle \frac{v}{1-\varepsilon}\int^{q_{1-\varepsilon}}_{v} \frac{k}{\sigma_{S}} \cdot g\left(\frac{\left(z-\mu_{S}\right)^{2}}{2 \sigma_{S}^{2}}\right) \mathrm{d} z &=& \displaystyle \frac{v}{1-\varepsilon}\int^{\frac{q_{1-\varepsilon}-\mu_\mathrm{S}}{\sigma_\mathrm{S}}}_{ \frac{v-\mu_\mathrm{S}}{\sigma_\mathrm{S}} } k \cdot g\left(\frac{z^2}{2}\right) \mathrm{d} z \\[4mm]
& = & \displaystyle \frac{v}{1-\varepsilon} \left( \mathrm{\Phi}\left(\frac{q_{1-\varepsilon}-\mu_\mathrm{S}}{\sigma_\mathrm{S}}\right)- \mathrm{\Phi} \left( \frac{v-\mu_\mathrm{S}}{\sigma_\mathrm{S}} \right) \right).
\end{array}
\end{equation*}
Hence, combine the constraint with (\ref{prob:optimistic chance constraint second constraint}), we have the following equivalent expression for problem~\eqref{prob:optimistic chance constraint 4}:
\begin{equation*}
\begin{array}{r@{\;\;}l@{\;\;}l}
\inf & v \\
{\rm s.t.}   & \displaystyle \int^{\frac{(q_{1-\varepsilon}-\mu_\mathrm{S})^2}{2\sigma_\mathrm{S}^2}}_{\frac{(v-\mu_\mathrm{S})^2}{2 \sigma_\mathrm{S}^2}}   k \cdot g(z) \mathrm{d} z 
+
\frac{\mu_\mathrm{S}-v}{\sigma_\mathrm{S}} \left(\mathrm{\Phi}\left(\frac{q_{1-\varepsilon}-\mu_\mathrm{S}}{\sigma_\mathrm{S}}\right) - \mathrm{\Phi}\left(\frac{v-\mu_\mathrm{S}}{\sigma_\mathrm{S}}\right) \right) \le \frac{\theta\Vert \bm{x} \Vert_{\bm{\Sigma}^{-1}}}{\sigma_\mathrm{S}}=\theta\\[3mm]
&   \displaystyle v \le \mathbb{P}_\mathrm{S}\text{-VaR}_{1-\varepsilon}[-\tilde{\bm{r}}^\top\bm{x}]\\
& v \in \mathbb{R},
\end{array}
\end{equation*}
where the equality follows from the definition of the Mahalanobis norm. Let $\eta = (v-\mu_\mathrm{S})/\sigma_\mathrm{S}$, the best-case VaR now becomes
\begin{equation}\label{prob:optimistic chance constraint 5}
\begin{array}{r@{\;\;}l@{\;\;}l}
\inf & \mu_\mathrm{S} + \sigma_\mathrm{S}\eta     \\
{\rm s.t.}   & \displaystyle \int^{(\mathrm{\Phi}^{-1}(1-\varepsilon))^2/2}_{\eta^2/2}   k \cdot g(z) \mathrm{d} z 
-
\eta \cdot(1-\varepsilon-\mathrm{\Phi}(\eta)) \le \theta \\[3mm]
&   \displaystyle \eta \le \mathrm{\Phi}^{-1}(1-\varepsilon)\\
& \eta \in \mathbb{R}.
\end{array}
\end{equation}
The function
\begin{equation*}
V(\eta) \triangleq \displaystyle \int^{(\mathrm{\Phi}^{-1}(1-\varepsilon))^2/2}_{\eta^2/2}   k \cdot g(z) \mathrm{d} z - \eta \cdot(1-\varepsilon-\mathrm{\Phi}(\eta))
\end{equation*}
is monotonically decreasing on $(-\infty, \mathrm{\Phi}^{-1}(1-\varepsilon))$ since for any $\eta < \mathrm{\Phi}^{-1}(1-\varepsilon)$, it holds that
\begin{equation*}
V'(\eta) = -\eta \cdot k \cdot g\left(\frac{\eta^2}{2}\right) - (1-\varepsilon) + \mathrm{\Phi}(\eta) + \eta\mathrm{\phi}(\eta) = \mathrm{\Phi}(\eta)-(1-\varepsilon) < 0.
\end{equation*}
Thus problem~\eqref{prob:optimistic chance constraint 5} can be efficiently solved be a bisection algorithm and the optimal $\underline{\eta}^\star$ as claimed can be obtained. Finally the result can be obtained as follows:
\begin{equation*}
\begin{array}{c@{\;\;}l}
\displaystyle \exists \; \mathbb{P} \in \mathcal{F}(\theta): \mathbb{P}[\tilde{\bm{r}}^\top\bm{x} \ge y] \ge 1-\varepsilon &  \displaystyle \Longleftrightarrow -y \ge \sigma_\mathrm{S}\underline{\eta}^\star + \mu_\mathrm{S} \\[3mm]
&\displaystyle \Longleftrightarrow \frac{-y-\mu_\mathrm{S}}{\sigma_\mathrm{S}} \ge \underline{\eta}^\star\\[3mm]
&\displaystyle \Longleftrightarrow \mathrm{\Phi}\left(\frac{-y-\mu_\mathrm{S}}{\sigma_\mathrm{S}}\right) \ge \mathrm{\Phi}(\underline{\eta}^\star)\\[3mm]
&\displaystyle \Longleftrightarrow \mathbb{P}_{(\bm{\mu},\bm{\Sigma},g)}\left[\frac{\tilde{\bm{r}}^\top\bm{x}-\mu_\mathrm{S}}{\sigma_\mathrm{S}} \ge \frac{y-\mu_\mathrm{S}}{\sigma_\mathrm{S}} \right] \ge 1-\bar{\varepsilon}\\[3mm]
&\displaystyle \Longleftrightarrow \mathbb{P}_{(\bm{\mu},\bm{\Sigma},g)}
[\tilde{\bm{r}}^\top\bm{x} \ge y] \ge 1-\bar{\varepsilon}.
\end{array}
\end{equation*}

\hfill\qed

With $\bar{\varepsilon}$ in Lemma~\ref{lemma:optimistic robust chance constraint}, we are now ready to derive a second-order cone reformulation of the distributionally optimistic chance-constrained model~\eqref{prob:optimistic chance constrained MDP}.

\begin{proposition}\label{prop:optimistic chance constraint socp}
Suppose in the Wasserstein ambiguity set \eqref{set:Wasserstein ball}, the reference distribution is an elliptical distribution $\hat{\mathbb{P}} = \mathbb{P}_{(\bm{\mu}, \bm{\Sigma}, g)}$ and the Wasserstein distance is equipped with a Mahalanobis norm associated with the positive definite matrix $\bm{\Sigma}$. If the risk threshold satisfies $\varepsilon \le \bar{\varepsilon} < 0.5$, then the distributionally optimistic chance-constrained MDP~\eqref{prob:optimistic chance constrained MDP} is equivalent to the second-order cone program
\begin{equation*}
\displaystyle \ell_{\rm DOCC}(\theta, \varepsilon) =  \max_{\bm{x} \in \mathcal{X}} \; \bm{\mu}^\top\bm{x} - \Vert{\rm \Phi}^{-1}(1-\bar{\varepsilon}) \bm{\Sigma}^{1/2}\bm{x} \Vert_2,
\end{equation*}
where $\bar{\varepsilon}=1-\mathrm{\Phi}(\underline{\eta}^\star)\ge\varepsilon$ with $\underline{\eta}^\star$ being the smallest $\eta \le \mathrm{\Phi}^{-1}(1-\varepsilon)$ that satisfies
\begin{equation*}
\eta({\rm \Phi}(\eta)-(1-\varepsilon))+\int_{\eta^{2} / 2}^{\left({\rm \Phi}^{-1}(1-\varepsilon)\right)^{2} / 2} k g(z) \mathrm{d} z \le \theta.
\end{equation*}
\end{proposition}

\noindent \textit{Proof.}
By Lemma~\ref{lemma:optimistic robust chance constraint}, the first constraint in \eqref{prob:optimistic chance constrained MDP} is equivalent to
\begin{equation*}
\mathbb{P}_{(\bm{\mu}, \bm{\Sigma}, g)}[\tilde{\bm{r}}^\top\bm{x}\ge y] \ge 1 - \bar{\varepsilon},
\end{equation*}
where $\bar{\varepsilon}=1-\mathrm{\Phi}(\underline{\eta}^\star)\ge\varepsilon$ with $\underline{\eta}^\star$ being the smallest $\eta \le \mathrm{\Phi}^{-1}(1-\varepsilon)$ that satisfies
\begin{equation*}
\eta({\rm \Phi}(\eta)-(1-\varepsilon))+\int_{\eta^{2} / 2}^{\left({\rm \Phi}^{-1}(1-\varepsilon)\right)^{2} / 2} k g(z) \mathrm{d} z \le \theta,
\end{equation*}
which can be further transformed as follows:
\begin{equation*}
\begin{array}{r@{\;\;}c@{\;\;}l}
\mathbb{P}_{(\bm{\mu},\bm{\Sigma},g)}[\tilde{\bm{r}}^\top\bm{x} \ge y] \ge 1-\bar{\varepsilon} & \Longleftrightarrow & {\rm \Phi}((\bm{\mu}^\top\bm{x}-y)/\sqrt{\bm{x}^\top\bm{\Sigma}\bm{x}}) \ge 1-\bar{\varepsilon}\\
& \Longleftrightarrow & \bm{\mu}^\top\bm{x}-y \ge {\rm \Phi}^{-1}(1-\bar{\varepsilon})\sqrt{\bm{x}^\top\bm{\Sigma}\bm{x}}\\
&\Longleftrightarrow & \bm{\mu}^\top\bm{x} - y \ge  \Vert{\rm \Phi}^{-1}(1-\bar{\varepsilon}) \bm{\Sigma}^{1/2}\bm{x} \Vert_2,
\end{array}
\end{equation*}
where the first equivalence holds by the linearity of elliptical distributions, the second one holds because of the non-decreasing cumulative distribution function $\mathrm{\Phi}(\cdot)$, and the third one holds as $\bar{\varepsilon} < 0.5$. Since the optimal value is achieved with $y = \bm{\mu}^\top\bm{x} - \Vert{\rm \Phi}^{-1}(1-\bar{\varepsilon}) \bm{\Sigma}^{1/2}\bm{x} \Vert_2$, plugging this equation in the objective of \eqref{prob:optimistic chance constrained MDP} then concludes our proof.
\hfill\qed

\section{Additional Details on Robust MDPs}\label{apd:RMDP}

As introduced in \cite{delage2010percentile}, robust MDPs maximizes the total expected return considering the worst-case realization of
the uncertain parameter within a predefined ambiguity set:
\begin{equation}\label{prob:rmdp}
\displaystyle\max_{\bm{\pi}\in\Pi}\min_{r^0\in\mathcal{R}, r^1\in\mathcal{R},\cdots}\mathbb{E}\left[\sum_{t=0}^{\infty}\gamma^t r^t(s_t)\;\vert\;s_0\propto \bm{p}_0, \bm{\pi}\right],
\end{equation}
where $\Pi$ is the set of all the stationary randomized policies, $r^t$ and $s_t$ are the reward and state at time stage $t$, respectively. As in \cite{delage2010percentile}, we set $\mathcal{R}$ to be the 99\% confidence ellipsoid of the random reward vector as the uncertainty set.

\section{Additional Details on BROIL}\label{apd:BROIL}
Similar to our return-risk model, BROIL \citep{brown2020bayesian} also seeks a policy that maximizes the weighted average of the mean and percentile performances: 
\begin{equation}\label{prob:broil}
\max_{\bm{\pi}\in\Pi} \lambda\cdot\mathbb{E}\left[\sum_{t=0}^{\infty}\gamma^t r^t(s_t)\;\vert\;s_0\propto \bm{p}_0, \bm{\pi}\right] + (1-\lambda)\cdot \mathrm{CVaR}_{\varepsilon}\left[\sum_{t=0}^{\infty}\gamma^t r^t(s_t)\;\vert\;s_0\propto \bm{p}_0. \bm{\pi}\right],
\end{equation}
where $\lambda\in[0,1]$ is the weight. Given $\bm{R}\in\mathbb{R}^{SA\times n}$ as the matrix of ($n$) reward samples,  BROIL can be expressed as a linear program as follows:
\begin{equation*}
\max_{\bm{x}\in\mathcal{X},y\in\mathbb{R}} \lambda\cdot\frac{1}{n}\bm{e}^\top\bm{R}^\top\bm{x} + (1-\lambda)\cdot\left(y-\frac{1}{\varepsilon}\cdot\frac{1}{n}\bm{e}^\top (y\cdot\bm{e}-\bm{R}^\top\bm{x})\right).
\end{equation*}
Observe that, there are two major differences between BROIL and our return-risk model: first, BROIL use CVaR as its risk measure, while VaR is applied in our return-risk model; second, while distributionally robustness is considered in (both the mean and VaR of return in) our objective function, BROIL only computes the nominal mean and CVaR of the return.

\section{Additional Details and Results on the Experiments}

\subsection{Additional Details of Parameter Selection}\label{apd:parameter}
We use cross validation for parameter selection in both the simulation and empirical studies. For DRMDPs~\eqref{prob:pessimistic MDP}, the candidate set for $\theta$ is $\{0,2,\cdots,18\}$; for CC~\eqref{prob:chance constrained MDP}, the candidate set for $\varepsilon$ is $\{i\varepsilon'/5\}_{i\in[5]}$; for RR~\eqref{RR}, we select $\theta$ such that $\underline{\varepsilon}$ varies among $\{i\varepsilon'/5\}_{i\in[5]}$, and we select $\alpha\in\{0,0.25,0.5,0.75,1\}$; for BROIL~\eqref{prob:broil}, we select $\lambda\times\varepsilon\in\{0,0.25,0.5,0.75,1\}\times\{0.05,0.1,0.15\}$; for RMDPs~\eqref{prob:rmdp}, as in \cite{delage2010percentile}, we set $\mathcal{R}$ to be the 99\% confidence ellipsoid of the random reward vector as the uncertainty set.

\subsection{Additional Details of the Simulation Study}\label{apd:simulation}
We consider $S = 10$ states, $A=10$ actions, a uniform initial state distribution, and a discount factor $\gamma=0.95$. For each state $s \in [S]$, the number of reachable next-state is $\lceil \log S \rceil$. We sample the true reward from a multivariate normal distribution $\mathcal{N}(\bm{\mu}', \bm{\Sigma}')$, where for each $k \in [SA]$, $\mu'_k$ is generated as follows: first we sample a number (0 or 1) from a discrete uniform distribution in $\{0,1\}$. If the result is 0, we generate $\mu'_k$ from the normal distribution $\mathcal{N}(50,100)$; otherwise we generate it from $\mathcal{N}(90,100)$. Standard deviations of rewards are generated in the same manner with another two normal distributions $\mathcal{N}(3,9)$ and $\mathcal{N}(18,9)$. Both standard deviations and means are trimmed to be non-negative after the above procedure.
The correlation matrix of rewards is generated as follows: we first sample a matrix $ \bm{R} \in \mathbb{R}^{SA \times SA} $ with all its entries independently sampled in $ [0.25,1] $ uniformly, and then obtain our correlation matrix ${\rm diag}(\bm{d})\bm{V}{\rm diag}(\bm{d})$, where $\bm{V} = \bm{R}^\top\bm{R} $ and $ \bm{d} = \{d_i\}_{i\in[SA]} = \{1/\sqrt{V_{ii}}\}_{i\in[SA]}$.

\subsection{Additional Details of the Empirical Study}\label{apd:empirical}
\begin{figure}[t]
\begin{center}
\includegraphics[width=1.0\textwidth]{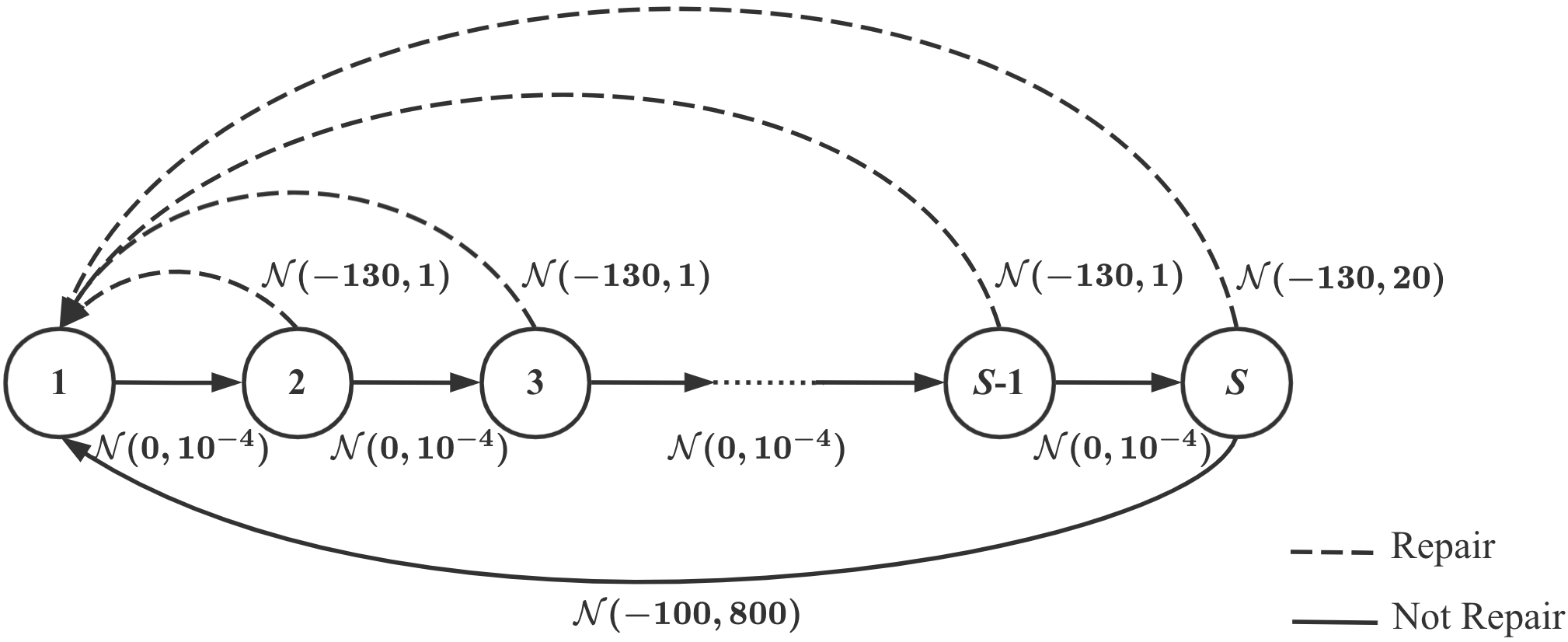}
\end{center}
\caption{\textnormal{A machine replacement problem with fixed Gaussian rewards.}}
\label{fig:machine replacement}
\end{figure}
In this experiment, each machine is subject to the same underlying MDP with a state set $\mathcal{S}= [S]$ with $S=50$ and an action set with only two actions: repair the machine or not. The transition is deterministic and the discount factor is $0.8$. The reward depends on both the current state and action, and all the rewards are independently and normally distributed. Figure~\ref{fig:machine replacement} illustrates the true underlying distribution that generates the random rewards. 

\subsection{Additional Results of the Simulation Study}\label{apd:simulation results}

\begin{figure}[t]
\begin{minipage}[t]{0.5\linewidth}
\centering
\includegraphics[width=3in]{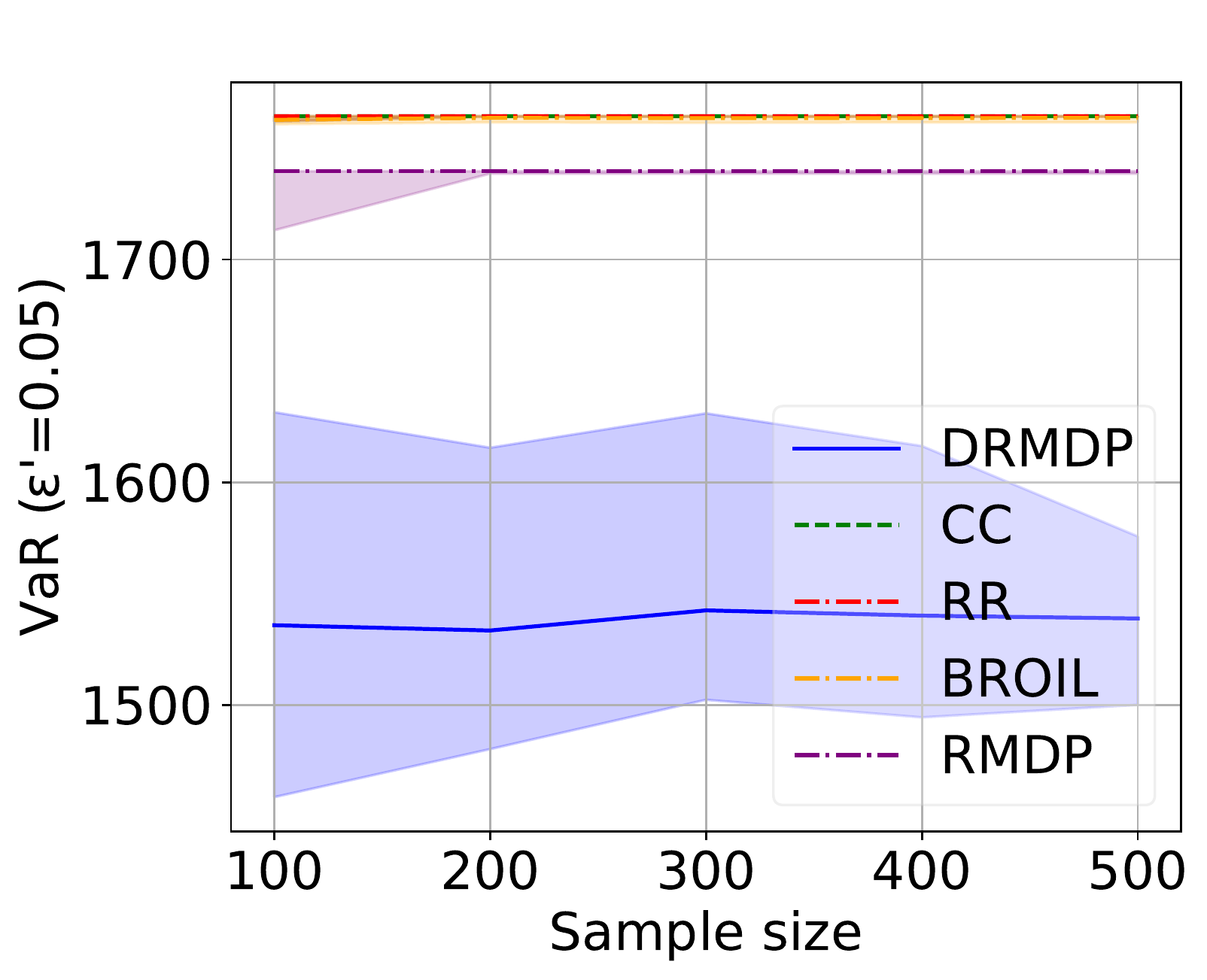}
\end{minipage}%
\begin{minipage}[t]{0.5\linewidth}
\centering
\includegraphics[width=3in]{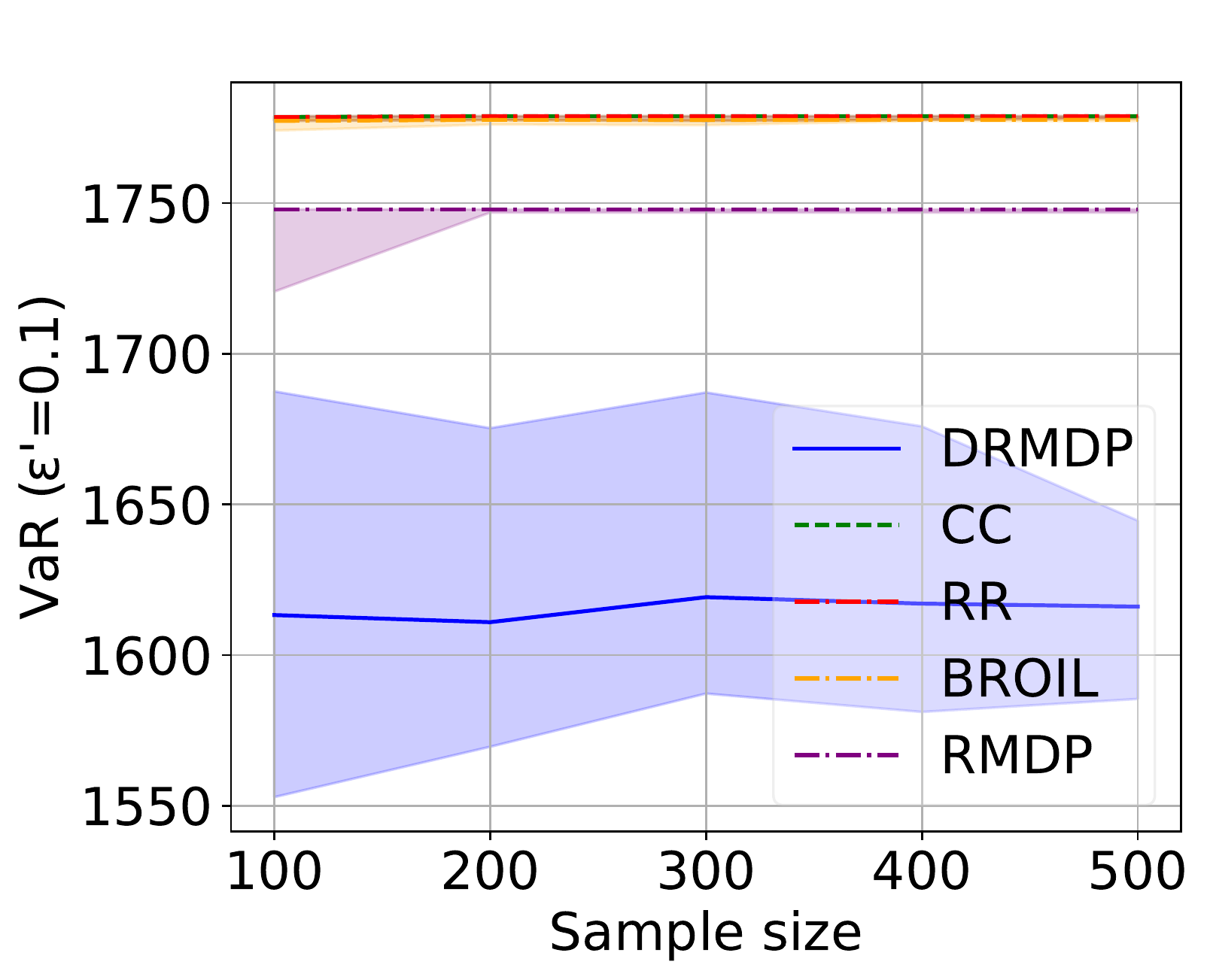}
\end{minipage}
\caption{\textnormal{Simulation. Models DRMDP~\eqref{prob:pessimistic MDP}, CC~\eqref{prob:chance constrained MDP}, RR~\eqref{RR}, RMDP and BROIL evaluated by VaR (risk threshold $\varepsilon'\in\{5\%,10\%\}$). The upper and lower edges of the shaded areas are respectively the 95\% and 5\% percentiles of the 100 performances, while the solid lines are the medians.}}
\label{fig:simulation 05 10}
\end{figure}

\subsection{Additional Results of the Empirical Study}\label{apd:empirical results}

\begin{figure}[h]
\begin{minipage}[t]{0.5\linewidth}
\centering
\includegraphics[width=3in]{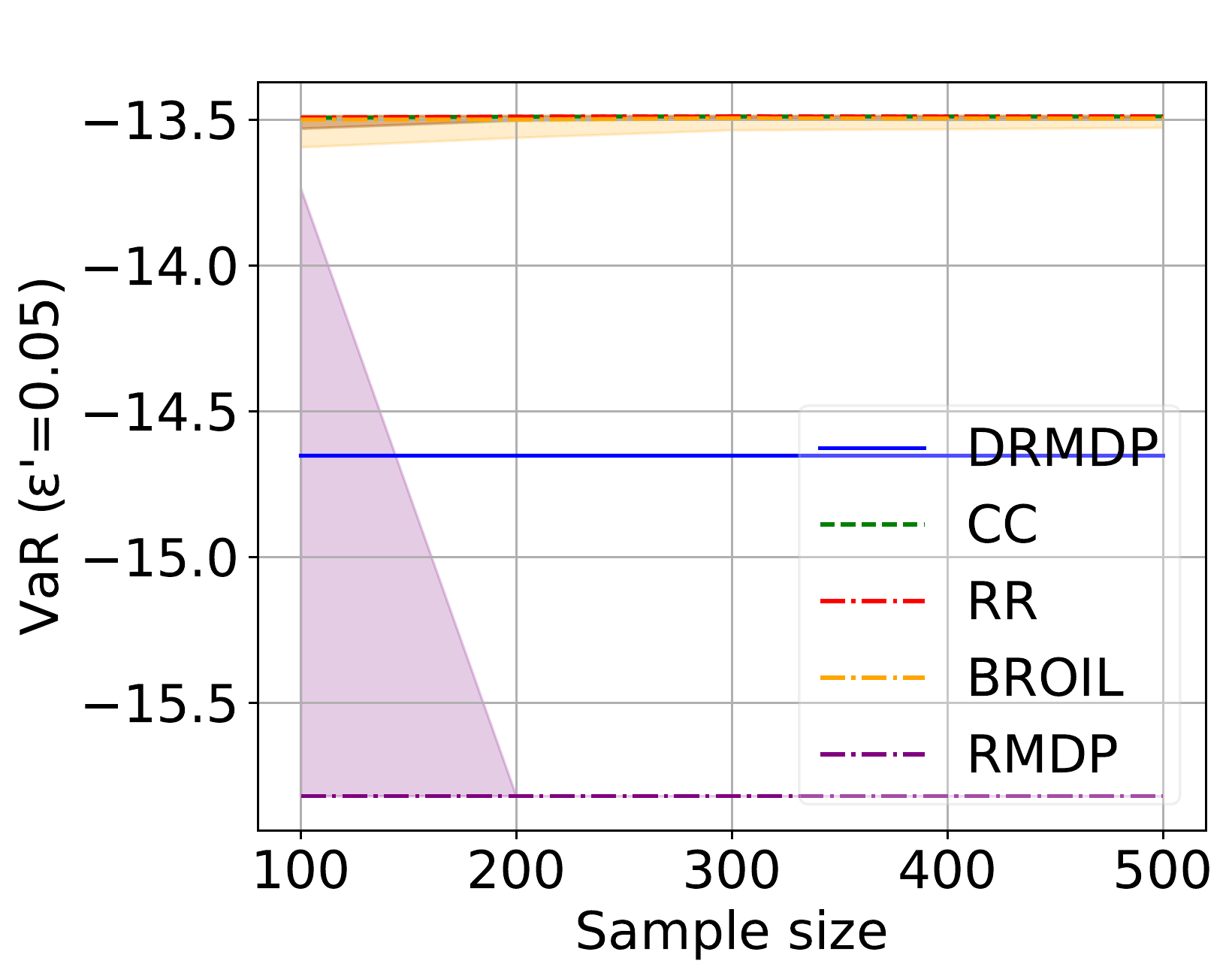}
\end{minipage}%
\begin{minipage}[t]{0.5\linewidth}
\centering
\includegraphics[width=3in]{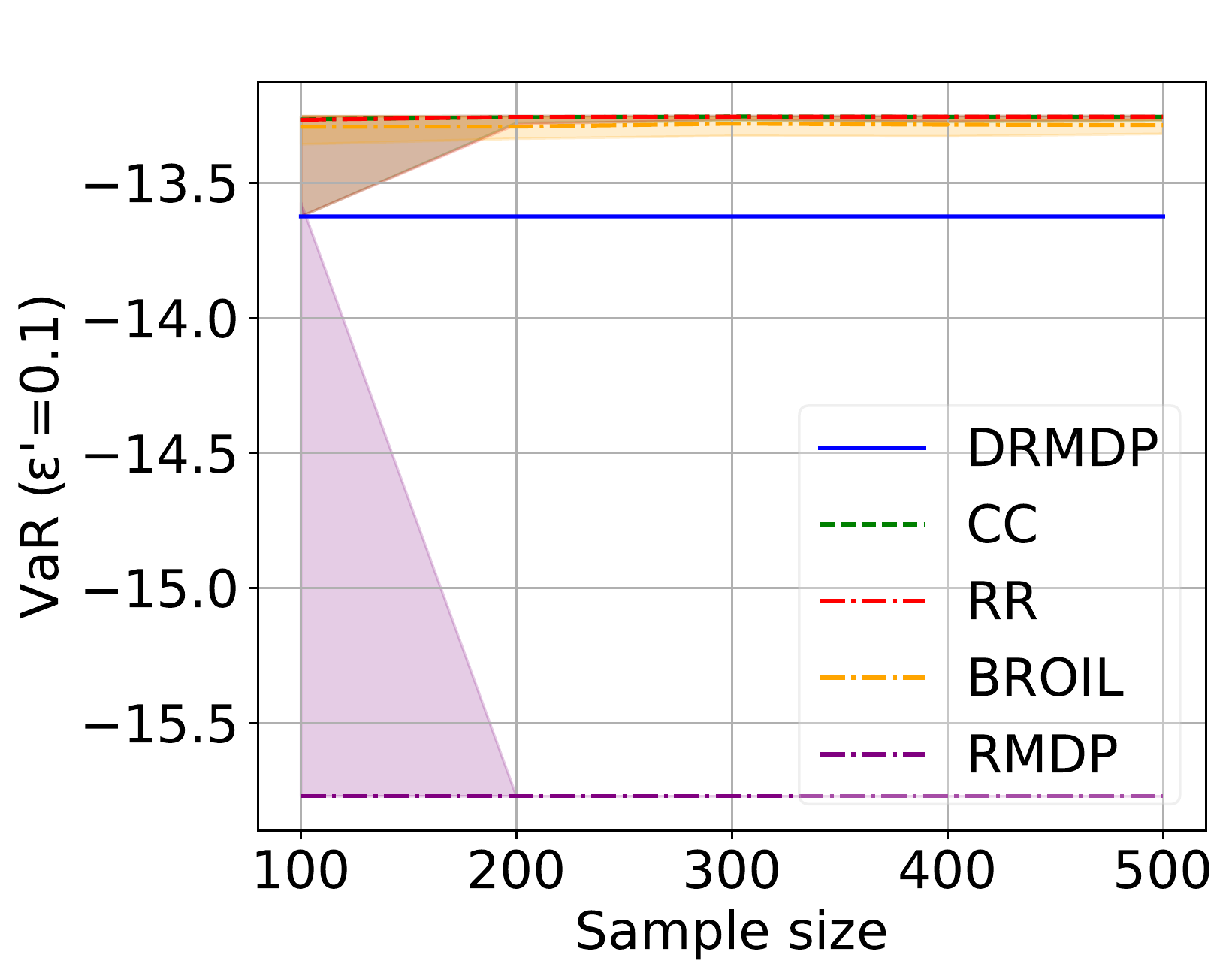}
\end{minipage}
\caption{\textnormal{Empirical. Models DRMDP~\eqref{prob:pessimistic MDP}, CC~\eqref{prob:chance constrained MDP}, RR~\eqref{RR}, RMDP and BROIL evaluated by VaR (risk threshold $\varepsilon'\in\{5\%,10\%\}$). The upper and lower edges of the shaded areas are respectively the 95\% and 5\% percentiles of the 100 performances, while the solid lines are the medians.}}
\label{fig:empirical 05 10}
\end{figure}

\newpage
\section{Related Works}\label{apd:related works}
Table~\ref{table:related works} summarizes literature that is related to our work. We remark that, compared to its related works in Table~\ref{table:related works}, our return-risk model is the only one that considers risk ambiguity, and we have also designed a fast first-order algorithm to obtain its solution, which enhance the practicality of our model for large-scale problems.

\begin{table}[h]
\caption{\textnormal{Related works.}}
\centering
\begin{tabular}{cccccc}
\toprule
Paper&Uncertainty & Robustness &Ambiguity set& Risk measure &  Soft-robustness  \\ \midrule
\cite{delage2010percentile} &\tabincell{c}{Rewards\\ and \\ transition kernel}&   - &  - &VaR       &  No      \\
\cite{xu2010distributionally} &\tabincell{c}{Rewards\\ and \\ transition kernel}&   DRO &Nested   &-       &  No      \\
\cite{yu2015distributionally} &\tabincell{c}{Rewards\\ and \\ transition kernel}&   DRO & (General) Nested  &-       &  No      \\
\cite{brown2020bayesian}&Rewards       &   -      &  - &CVaR   & Yes     \\
\cite{gilbert2017optimizing}&Rewards      & -     &   -  &VaR      & No   \\
\cite{lobo2020soft} & Transition kernel     &   -      & -  &CVaR       & Yes     \\
\cite{yang2020wasserstein} & Transition kernel     &   DRO      & Wasserstein   &-       & No     \\
This paper &Rewards      &   DRO      &  Wasserstein &VaR       & Yes     \\
\bottomrule 
\end{tabular} \label{table:related works}
\end{table}


\end{appendices}

\end{document}